\documentclass{article}

\usepackage[preprint]{neurips_2022}

\usepackage{amsmath,amsfonts,bm}

\def\figref#1{figure~\ref{#1}}

\def\eqref#1{equation~\ref{#1}}

\def\1{\bm{1}}

\DeclareMathAlphabet{\mathsfit}{\encodingdefault}{\sfdefault}{m}{sl}
\SetMathAlphabet{\mathsfit}{bold}{\encodingdefault}{\sfdefault}{bx}{n}

\newcommand{\E}{\mathbb{E}}

\newcommand{\R}{\mathbb{R}}

\usepackage[utf8]{inputenc} 
\usepackage[T1]{fontenc}    
\usepackage{amsfonts}       
\usepackage{nicefrac}       
\usepackage{xcolor}         
\usepackage{microtype}
\usepackage{url}
\usepackage{graphicx}
\usepackage{multirow}
\usepackage{booktabs}
\usepackage{subcaption}
\usepackage{wrapfig}

\newcommand{\D}{\mathcal{D}}

\usepackage{mathtools}
\newcommand{\underdescribe}[3][0pt]{\hspace*{.12em}\underbracket[0.5pt][1pt]{#2\hspace*{#1}}_{#3}}

\title{An Investigation of the Bias-Variance\\Tradeoff in Meta-Gradients}

\author{%
  Risto Vuorio \thanks{contact: \texttt{risto.vuorio@cs.ox.ac.uk}}\\
  University of Oxford \\
  \And
  Jacob Beck \\
  University of Oxford \\
  \And
  Shimon Whiteson \\
  University of Oxford \\
  \AND
  Jakob Foerster \\
  University of Oxford \\
  \And
  Gregory Farquhar \\
  DeepMind
}

\begin{document}

\maketitle

\begin{abstract}
  Meta-gradients provide a general approach for optimizing the meta-parameters of reinforcement learning (RL) algorithms.
  Estimation of meta-gradients is central to the performance of these meta-algorithms, and has been studied in the setting of MAML-style short-horizon meta-RL problems.
  In this context, prior work has investigated the estimation of the Hessian of the RL objective, as well as tackling the problem of credit assignment to pre-adaptation behavior by making a \emph{sampling correction}.
  However, we show that Hessian estimation, implemented for example by DiCE and its variants, always adds bias and can also add variance to meta-gradient estimation.
  Meanwhile, meta-gradient estimation has been studied less in the important long-horizon setting, where backpropagation through the full inner optimization trajectories is not feasible.
  We study the bias and variance tradeoff arising from truncated backpropagation and sampling correction, and additionally compare to evolution strategies, which is a recently popular alternative strategy to long-horizon meta-learning.
  While prior work implicitly chooses points in this bias-variance space, we disentangle the sources of bias and variance and present an empirical study that relates existing estimators to each other.
\end{abstract}

\section{Introduction}
Recently, meta-gradients have been used to adapt the hyperparameters of state-of-the-art policy gradient RL algorithms online \citep{zahavy2020self}, as well as for learning black box RL algorithms from scratch \citep{oh2020discovering}.
The estimation of meta-gradients is central to the performance of these algorithms, but has received limited focused study.
Deriving unbiased estimators of meta-gradients requires care because the meta-parameters induce changing parameters, affecting the data distribution used to compute the meta-gradient.
\citet{al2017continuous} present an unbiased meta-gradient estimator that correctly accounts for the changes in the data distribution by adding a \textit{sampling correction} to the direct meta-gradient.
However, their derivation and experiments are limited to meta-RL with short optimization trajectories in the inner loop.
Further work studying meta-gradient estimation for meta-RL is also largely restricted to this context.

Computing the meta-gradient requires differentiating through the agent's parameter update, which may itself include an estimated policy gradient.
\citet{foerster2018dice} and others
have claimed that an estimate of the \textit{expected} policy Hessian should be included in the meta-gradient estimator, and provide methods for calculating such an estimate or reducing its variance.
In this work, we present the surprising result that using an estimate of the expected Hessian in the meta-gradient estimator actually \emph{adds} bias, which can degrade performance.
Such a term would only make sense if the agent made an \textit{expected} policy gradient update, rather than the sampled update that must be taken using finite data.
Instead, the correct meta-gradient of a standard policy gradient update can be computed by straightforward backpropagation of a surrogate loss, as is typical in RL.
This focus on estimating expected Hessians may have distracted the community from the challenges in meta-gradient estimation in the long-horizon setting.


In the short-horizon setting, the meta-gradient can be computed via backpropagation over the whole optimization trajectory of the inner learning problem.
Due to computational constraints, this is not feasible in the long-horizon setting.
A common approach is to compute the meta-gradient over a truncated optimization horizon.
As a result of truncation, neither the direct meta-gradient nor the sampling correction may be calculated exactly.
It is therefore unclear how the truncated sampling correction affects the bias and variance of the meta-gradient estimation.
In this paper, we study how bias and variance may be traded off by varying the truncation horizon and the weight given to the sampling correction.
We find that, while sampling correction does not uniformly reduce the bias of truncated estimators, it nevertheless yields convergence to better local optima.
The cost of using the sampling correction is increased variance.

Another approach to meta-learning over long horizons is to use black box methods such as evolution strategies (ES)~\citep{rechenberg1973evolutionsstrategie,schwefel1977evolutionsstrategien}.
On one hand, as a black box method, ES does not require new derivations for handling the changing data distribution and it has favorable memory requirements when used with long optimization horizons.
On the other, its sample complexity grows with the number of parameters being optimized.
We empirically relate the bias and variance of backpropagation based meta-gradient estimators to those of ES and find a cross-over point, where the variance of the sampling corrected meta-gradient becomes higher than that of ES.
We also compare the different estimators in a long-horizon setting in practice and find that none of them are good enough for training complex meta-parametrizations.






Our contributions are as follows.
(1) We show mathematically why using the expected policy Hessian estimator is incorrect and demonstrate that the resulting bias harms performance in practice.
(2) We show how the variance of the sampling correction for meta-gradients grows counterintuitively, how it works in the truncated optimization setting, and how standard advantage estimation fails for the sampling correction.
(3) We characterize the bias-variance tradeoff due to truncated optimization and the sampling correction in an empirical study, relating existing approaches to each other and showing how to interpolate between them. We also compare the backpropagation-based estimators to ES.

\section{Background}
\label{sec:background}

We assume that the RL algorithm is learning a control policy in a Markov decision process (MDP) defined by a tuple $(\mathcal{S}, \mathcal{A}, p_0, p, r, \gamma)$, where $\mathcal{S}$ is the space of states, $\mathcal{A}$ the space of actions, $p_0(s_0)$ the distribution over initial states, $p(s_{t+1} | s_t, a_t)$ the conditional probability distribution of the next state given a state and an action, $r(s_t, a_t)$ the reward function for the transition, and $\gamma \in [0, 1)$
a discount factor.
A trajectory $\tau = (s_0, a_0, s_1, a_1, \dots, s_H, a_H) \in \mathcal{T}$, is a sequence of states and actions sampled from the dynamics defined by the MDP and the policy, where the horizon $H$ may be infinite. $\mathcal{T}$ is the space of trajectories and
$\pi(a_t|s_t;\theta)$ is the policy parametrized by $\theta \in \R^{d_{\theta}}$.
The objective of an RL algorithm is to train a policy that maximizes the expected discounted sum of rewards: $J(\theta) = \E_{\tau \sim p(\tau; \theta)} \left[\sum_{t=0}^{H} \gamma^t r(s_t, a_t)\right]$. 
For brevity we
write $R(\tau) = \sum_{t=0}^{H} \gamma^t r(s_t, a_t)$, where $s_t$ and $a_t$ come from $\tau$.
%
Policy gradient algorithms update $\theta$ by gradient ascent on the policy gradient:
\begin{align}
    \nabla_{\theta} J(\theta) = \E_{\tau \sim p(\tau; \theta)} \left[\sum_{t=0}^{H} \nabla_{\theta} \log \pi(a_t | s_t; \theta) R(\tau)\right]. \label{eq:expected_policy_gradient}
\end{align}
Typically, the expectation in \eqref{eq:expected_policy_gradient} cannot be evaluated exactly and is instead approximated from samples by
\begin{equation}
    \nabla_{\theta} J(\theta, \D) = \frac{1}{|\D|}  \sum_{\tau \in \D} \sum_{t=0}^{H} \nabla_{\theta} \log \pi(a_t | s_t ; \theta) R(\tau) \label{eq:stochastic_policy_gradient},
\end{equation}
where $\D \in \mathcal{T}^N$ is a batch of $N$ trajectories collected with the policy parametrized by $\theta$.
The probability of the batch is $p(\D ; \theta) = \prod_{\tau \in \D} p(\tau ; \theta)$.

Meta-gradients arise in the meta-learning problem of optimizing the parameters $\eta$ of an update function $\Psi(\eta, \theta, \D): \R^{d_{\eta}} \times \R^{d_{\theta}} \times \mathcal{T}^N \rightarrow \R^{d_{\theta}}$,
which computes an update to the parameters for a policy.
For example, a simple update function is $\theta^{i+1} = \theta^i + \Psi_{lr}(\eta, \theta^i, \D^i) = \theta^i + \eta \nabla_{\theta^i} J(\theta^i, \D^i)$,
where the meta-parameter is the learning rate of a stochastic policy gradient update.
The index $i$ orders the policy parameters in a sequence of corresponding policy updates produced using $\Psi$.
In general, the meta-parameters can appear in any term in $\Psi$.
Meta-gradients are often computed in a setting where the policy is updated $K$ times using the update function and the meta-parameters are optimized to maximize the return of each of the $K+1$ policies.
We refer to $K$ as the lifetime of the agent.
The objective for such $K$-step meta-gradient is given by
\begin{align}
    J_{K}(\eta) = \sum_{k=0}^{K} \E_{\substack{\{\D^i \sim p(\D^i ; \theta^{i})\}_{i=0}^{k-1}}} \bigg[ \E_{\tau \sim p(\tau ; \theta^k)} \bigg[R(\tau)\bigg]\bigg], \label{eq:meta_objective}
\end{align}
where the outer expectation is taken over data sampled with all of the policies along the update trajectory so far, as each $\theta^k$ depends on all of the previous data through $\Psi$.
The objective in \eqref{eq:meta_objective} considers the value of all policies in the sequence of updates.
Some prior work, including MAML \citep{finn2017model}, instead uses a meta-objective which considers only the value of the final policy.
All of the derivations below are straightforward to apply to this meta-objective as well.

When $K$ is large, computing the exact meta-gradient with backpropagation is challenging due to limited memory and the rugged objective landscape \citep{metz2019understanding}.
Hence, backpropagation along the trajectory of updates is commonly truncated to a window of optimization steps much shorter than $K$.
Computing the meta-gradient over such truncation windows results in bias.
Nonetheless, this approach has been employed successfully in practical meta-RL algorithms, e.g.\ \citep{oh2020discovering}.
In addition to optimizing equation~\ref{eq:meta_objective}, truncated meta-gradient estimators are used for meta-learning online, alongside a normal RL algorithm.
In this online setting, instead of searching for the best stationary solution, the meta-learning ``tracks'' the inner loop learning and updates the meta-parameters continuously~\citep{sutton2007role,xu2018meta,zahavy2020self}.

\subsection{Unbiased meta-gradients}
\citet{al2017continuous} derived a general unbiased meta-gradient estimator for a $K$-step sequence of updates; we present a similar derivation here for convenience.
For a sequence of $K$ parameter updates the $K$th
parameter is:
\begin{align}
    \theta^K = \theta^0 + \sum_{i=0}^{K-1} \Psi(\eta, \theta^i, \D^i). \label{eq:sequence_of_updates}
\end{align}
Because the $\D^i$ are random variables, $\theta^i$ for $i>0$ are random variables too.
They depend on the initial parameters $\theta^0$, the meta-parameter $\eta$, and the data.
The meta-gradient estimator for the objective in \eqref{eq:meta_objective} on the sequence of updates in \eqref{eq:sequence_of_updates} can be derived as follows:
\begin{align}
    \nabla_{\eta} J_K(\eta) &= \nabla_{\eta} \sum_{k=0}^{K} \E_{\substack{\{\D^i\}_{i=0}^{k-1}}} \bigg[ \E_{\tau \sim p(\tau ; \theta^k)} \left[R(\tau)\right]\bigg] \nonumber \\
    &\hspace{-4em}= \sum_{k=0}^{K} \E_{\substack{\{\D^i\}_{i=0}^{k-1} \\ \tau \sim p(\tau ; \theta^k)}} \bigg[ \bigg(\underdescribe{ \sum_{j=0}^{k - 1} \nabla_{\eta} \theta^j \nabla_{\theta^{j}} \log p(\D^j ; \theta^j)}{\texttt{sampling correction}}
    + \underdescribe{\vphantom{\sum_{j=0}^{k-1}}\nabla_{\eta} \theta^k \nabla_{\theta^{k}} \log p(\tau ; \theta^k)}{\texttt{direct meta-gradient}} \bigg) R(\tau)\bigg], \label{eq:unbiased_meta_grad}
\end{align}
where $\nabla_{\eta} \theta^k \in \R^{d_{\eta} \times d_{\theta}}$ is the derivative w.r.t.\ $\eta$ of the sequence of updates resulting in $\theta^k$.
See appendix \ref{appendix:meta_grad_derivation} for a more detailed derivation.
Following from the additive sequence of updates in \eqref{eq:sequence_of_updates}, $\nabla_{\eta} \theta^k$ further expands as
\begin{align}
    \nabla_{\eta} \theta^k =  \nabla_{\eta} \theta^0 + \sum_{i=0}^{k-1} \bigg(&\nabla_{\eta} \Psi(\eta, \theta^i, \D^i)
    + \nabla_{\eta} \theta^i \nabla_{\theta^i}\Psi(\eta, \theta^i, \D^i)  \bigg), \label{eq:meta_jacobian}
\end{align}
where $\nabla_{\eta}\theta^0$ can be nonzero when the initial parameters are meta-learned, as in MAML.

The $\nabla_{\eta} \theta^k \nabla_{\theta^{k}} \log p(\tau ; \theta^k)$ terms in \eqref{eq:unbiased_meta_grad} are the product of the derivative of the $k$th policy parameter w.r.t.\ the meta-parameters and the standard policy gradient.
The terms of the sum $\sum_{j=0}^{k - 1} \nabla_{\eta} \theta^j \nabla_{\theta^{j}} \log p(\D^j ; \theta^j)$ give the sampling correction that assigns credit from the experience collected with the updated policy directly to the earlier policies.
\citet{al2017continuous} show that these latter terms are missing from the original MAML derivation \citep{finn2017model}.
They are also omitted in most other meta-gradient algorithms \citep{xu2018meta,oh2020discovering,zheng2018learning,flennerhag2021bootstrapped}.
We discuss the original MAML derivation in appendix \ref{appendix:expected_meta_gradient}.

\subsection{Estimating the Hessian of policy value}
\label{subsec:value_hessian}

The standard approach for estimating the first derivative of $J(\theta)$ is to construct a surrogate loss of the form $\sum_{\tau \in \D}\sum_{s, a \in \tau}\log \pi(a | s ;\theta) R(\tau)$ and differentiate it w.r.t.\ $\theta$.
\citet{foerster2018dice} show that naively differentiating this surrogate loss a second time does not compute an unbiased estimate of the Hessian $\nabla^2_\theta J(\theta)$.
They propose a new surrogate objective called DiCE that may be differentiated any number of times to compute estimators of any-order derivatives of $J(\theta)$.
Subsequent work \citep{rothfuss2018promp,liu2019taming,farquhar2019loaded,mao2019baseline,tang2021unifying} builds on this approach or proposes alternative strategies for estimating this Hessian, which is assumed to be important for meta-gradient estimation.
However, contrary to the claims made by \citet{foerster2018dice} and others, we show below that when estimating the Hessian for meta-gradients in practice, using an estimate of $\nabla^2_\theta J(\theta)$ always adds bias and has high variance whereas the naive approach does not add bias.

\subsection{Evolution strategies for meta-gradient estimation}
\label{subsec:blackbox}
Evolution strategies (ES)~\citep{rechenberg1973evolutionsstrategie,schwefel1977evolutionsstrategien} have recently become a popular alternative for truncated backpropagation in meta-learning~\citep{kirsch2021introducing,vicol2021unbiased}.
In this paper we consider an ES algorithm by \citet{salimans2017evolution}.
In the following, we refer to this algorithm as just ES.
Consider a parameterised distribution $p(\theta; \psi)$ over the parameters $\theta$.
Then, ES conducts gradient ascent on the score function estimator of the expected value of some objective $F(\theta)$ over this distribution: $\E_{\theta \sim p(\theta;\psi)}\left[ \nabla_{\psi} \log p(\theta ; \psi ) F(\theta)\right]$.
This corresponds to optimizing a smoothed version of the true objective $F(\theta)$.
\citet{metz2019understanding} shows that the objective surface for meta-learning with many inner-loop updates is not smooth, and the meta-gradient estimate given by ES has lower variance than the backpropagation-based one due to this smoothing.
The drawback of ES is that its sample complexity scales linearly with the dimension of $\theta$~\citep{nesterov2017random}.

\section{Bias and variance of meta-gradient estimators}
\label{sec:theory}
In this section, we investigate the bias and variance of meta-gradients due to the Hessian estimation and sampling corrections in theory.
We first show the surprising result that including the Hessian of the \emph{expected} inner objective adds rather than removes bias.
Then, we explore three key considerations relating to the sampling correction terms, which are missing from most current meta-gradient algorithms and have not been discussed widely in the literature.
First, we show how the variance of the sampling correction terms grows counterintuitively with the batch size of the inner loop, and present
a weighting scheme for mitigating variance.
Second, we discuss the bias from sampling corrections in the truncated optimization setting.
Third, we show that standard advantage estimation with a state-dependent baseline results in bias when used with the sampling correction terms.

\subsection{Estimating the Hessian of the inner-loop objective}
\label{sec:expected_hessians}
If the inner-loop update function is gradient-based, then the term $\nabla_{\theta^k}\Psi(\eta, \theta^k, \D^k)$ in \eqref{eq:meta_jacobian} induces a second-order derivative: the Hessian of the objective that is differentiated in the inner-loop.
If this inner-loop objective was the expected policy value $J(\theta)$, then this Hessian is exactly the Hessian of policy value discussed in section \ref{subsec:value_hessian}:
\begin{align}
    \nabla_{\theta}^2 J(\theta)
    = \E_{\tau \sim p(\tau ; \theta)} \bigg[& \bigg( \nabla_{\theta} \log \pi(\tau ; \theta) \nabla_{\theta} \log \pi(\tau ; \theta)^{\top}
    +   \nabla_{\theta}^2 \log \pi(\tau ; \theta)   \bigg) R(\tau) \bigg]. \label{eq:expected_hessian}
\end{align}
The works discussed in section \ref{subsec:value_hessian} either directly estimate this quantity through sampling or propose lower-variance, but biased, estimators.  Regardless, they all refer to \eqref{eq:expected_hessian} as the desired quantity.

However, in almost all practical cases, the \emph{expected} policy gradient $\nabla_{\theta} J(\theta)$ is intractable and thus cannot be used in the inner loop.
Instead, we must use the gradient of the \emph{sampled} objective $\nabla_{\theta} J(\theta, \D)$ given by \eqref{eq:stochastic_policy_gradient}.
The update function is then a deterministic function of its inputs, which now include the data.
The Hessian of this sampled objective is consequently easy to compute with reverse mode automatic differentiation, simply by backpropagating through the computation graph of the sequence of updates in equation \ref{eq:sequence_of_updates}.
This is indeed what many meta-gradient implementations do in practice \citep{zheng2020can, zahavy2020self}.
Computing the Hessian of the sampled objective gives:
\begin{equation}
    \nabla_{\theta}^2 J(\theta, \D) = \frac{1}{|\D|}  \sum_{\tau \in \D} \nabla_{\theta}^2 \log \pi(\tau ; \theta) R(\tau).
\end{equation}
The term $\nabla_{\theta} \log \pi(\tau ; \theta) \nabla_{\theta} \log \pi(\tau ; \theta)^{\top} R(\tau) \in \R^{d_{\theta} \times d_{\theta}}$ does not appear in this stochastic Hessian.
Estimates of the Hessian of the expected objective thus include this spurious term which introduces bias, except in edge cases for which it vanishes, such as MDPs with no rewards.
Therefore, the meta-gradient estimators discussed in section \ref{subsec:value_hessian}, which estimate the Hessian by equation~\ref{eq:expected_hessian}, are biased in practice.
In section \ref{sec:experiments}, we illustrate the bias and variance resulting from using these estimators in practice.

\subsection{Variance due to the sampling correction}
\label{subsec:variance}
The sampling correction terms, of the form $\nabla_{\eta} \theta^j \nabla_{\theta^{j}} \log p(\D^j; \theta^j) R(\tau)$ where $\log p(\D^j; \theta^j) = \sum_{\tau \in \D^j} \log p(\tau ; \theta^j)$, require the gradient of the log probability of the entire batch of samples $\D^j$, and so involve a sum over the batch elements.

Intuitively, in normal policy gradient RL the probability of sampling a particular reward can only depend on the trajectory leading to that reward as the parameters $\theta$ are fixed.
In meta-gradient meta-RL, by contrast, the parameters $\theta^j$ vary, and depend on every trajectory used by the update function up to iteration $j$.
The sampling correction accounts for these dependencies.
Unfortunately, this means the variance of the sampling correction counterintuitively grows with the batch size of the inner loop.
The variance is of course still reduced by increasing the meta-batch size in the outer-loop.

In normal RL, although the dependencies are restricted to a single trajectory, a discount factor is still typically used to downweight the credit assigned to temporally distant actions.
This reduces variance, but introduces bias with respect to the undiscounted objective.
Similarly, we may downweight the sampling correction terms to trade off bias and variance in meta-gradient meta-RL.
One strategy for enabling this trade-off is to uniformly weight the sampling correction terms by a coefficient $\lambda$:
\begin{align}
    \sum_{k=0}^{K}
     \frac{1}{|\D|} \bigg(\lambda &\sum_{j=0}^{k - 1} \nabla_{\eta} \theta^j \nabla_{\theta^{j}} \log p(\D^j ; \theta^j)  \sum_{\tau \in \D^k} R(\tau)
      +  \sum_{\tau \in \D^k}  \nabla_{\eta} \theta^k \nabla_{\theta^{k}} \log p(\tau ; \theta^k) R(\tau) \bigg). \label{eq:sampled_meta_gradient}
\end{align}
This sampling correction coefficient is related to the $\lambda$ hyperparameter of E-MAML \citep{stadie2018some}, but we do not separately divide by the batch size, which is done in implementations of E-MAML though not motivated in the paper.
Therefore, the meta-gradient estimator is unbiased when our $\lambda=1.0$.
We investigate the bias-variance tradeoff induced by this $\lambda$ in section~\ref{sec:experiments}.
The coefficient $\lambda$ is only one possible scheme to trade off bias and variance.
Motivated by analogy to the usual discounting procedure in RL, we consider an exponential discounting in appendix \ref{appendix:exp_discount_meta_grad}.

\subsection{Sampling corrections in truncated optimization}
\label{subsec:truncated_optimization}
To the best of our knowledge none of the algorithms using truncated backpropagation employ any form of sampling correction, and it is not known how the sampling correction works in the truncated setting.
The truncation changes the terms in \eqref{eq:meta_jacobian}, making it a biased estimator.
Since the sampling correction terms are a product of the derivative in equation~\ref{eq:meta_jacobian} and the gradient of the log probability of a batch, they also become biased when truncation is used.
Because of the complex dynamics of the optimization process, we have no reason to believe that the bias from truncating the backpropagation decreases monotonically with the truncation length.
Therefore, it is possible that using the sampling correction on a truncated meta-gradient estimator increases rather than decreases the bias.
In addition to changing the derivatives, approximating the sum over $K$ in \eqref{eq:sampled_meta_gradient} with a sum over a shorter window causes bias by modifying the surrogate loss even before backpropagation.
We investigate the effect of sampling correction on the truncated estimators empirically in section \ref{sec:experiments}.

\subsection{Advantage estimation for sampling corrections}
\label{sec:advantage}
Intuitively, we would like to use a standard advantage estimator, such as GAE proposed by \citet{schulman2015high}, for the sampling correction terms in \eqref{eq:unbiased_meta_grad}.
However, we now show that using standard advantage estimation with the sampling correction results in bias, due to dependencies between the previous policies on the update trajectory and the value estimates of the updated policies.

The policy gradient via the surrogate loss with a simple advantage estimator is given by
$\nabla_{\theta} J(\theta) = \E_{s, a \sim p(s, a;\theta)} [\nabla_{\theta} \log \pi(a|s; \theta) (Q(s, a) - B(s))]$,
where $Q(s, a) = \E_{\tau \sim p(\tau; \theta)}[\sum_{t=0}^{\infty} \gamma^t r(s_t, a_t) | s_0 = s, a_0 = a]$ and $B(s)$ is a control variate, which can be for example the state-value given by $B(s) = V(s) = \E_a [Q(s, a)]$.
The simple advantage estimator $Q(s, a) - B(s)$ is unbiased because $s$ is independent of $a$ from which follows $\E_{a} [\nabla_{\theta} \log \pi(a|s; \theta)B(s)] = 0$.
In \eqref{eq:unbiased_meta_grad}, it is tempting to replace $R(\tau)$ with an advantage estimator given by $R(\tau) - \hat{V}(\tau)$, where $\hat{V}(\tau)$ is an estimate of the value of the policy in the first state of the trajectory $\tau$ and $\hat{V}$ is trained on the data $\D^i$ for $0 \leq i < k$.
This would allow sharing the value estimator between the inner-loop and outer-loop.
To get an unbiased gradient estimate, we need $\E_{\substack{\D^i,\, \tau \sim p(\tau ; \theta^k)}} [\nabla_{\eta} \log p(\D^i;\theta^i)\hat{V}(\tau)]$ for $0 \leq i < k$ to equal $0$.
However, this can be nonzero and thus lead to bias because the value function estimator $\hat{V}$ itself depends on the data $\D^i$.
More generally, any $B(s)$, which depends on $\D^i$ leads to bias.
In appendix \ref{sec:advantage_experiment} we demonstrate empirically the bias due to the dependence of the value estimator on $\D^i$.


\section{Experiments}
\label{sec:experiments}
In this section, we conduct experiments to test the bias-variance tradeoff in practice.
We first consider a bandit setting to illustrate the sources of bias and the bias-variance tradeoff in a simple case.
Then, to verify that the findings from the theory and the bandit experiments apply to the full RL problem, we consider meta-learning in a gridworld.

In the bandit setting, we train agents with REINFORCE~\citep{williams1992simple} for multiple update steps on bandits with randomly sampled arm rewards.
The meta-learning problem is to learn separate learning rates for the early and late parts of the inner learning.
Hence, these two meta-parameters define a learning rate schedule; intuitively, the agent must meta-learn an optimal trade-off of exploration and exploitation for a given distribution of bandit tasks.
The meta-parameters are optimized over multiple parallel lifetimes and updated after the lifetimes have concluded, at which point the agent is reset and new lifetimes are sampled.
The full details on the problem setting are given in appendix \ref{appendix:bandit_details}.
We run all experiments using JAX~\citep{jax2018github}, evosax~\citep{evosax2022github}, and Tune~\citep{liaw2018tune}.

\subsection{Sampling correction and DiCE in practice}
\label{subsec:sc_and_dice}
\begin{figure}
    \centering
    \begin{subfigure}[b]{0.48\textwidth}
      \centering
      \raisebox{12pt}{\includegraphics[width=\textwidth]{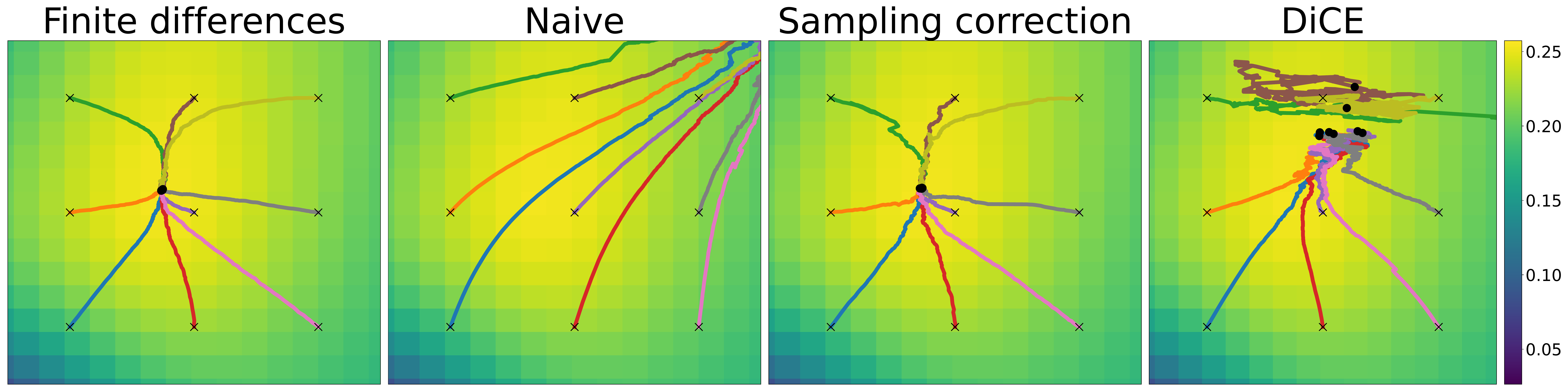}}%
      \caption{Meta-parameter trajectories}
    \end{subfigure}
    \quad
    \begin{subfigure}[b]{0.48\textwidth}
      \centering
      \includegraphics[width=\textwidth]{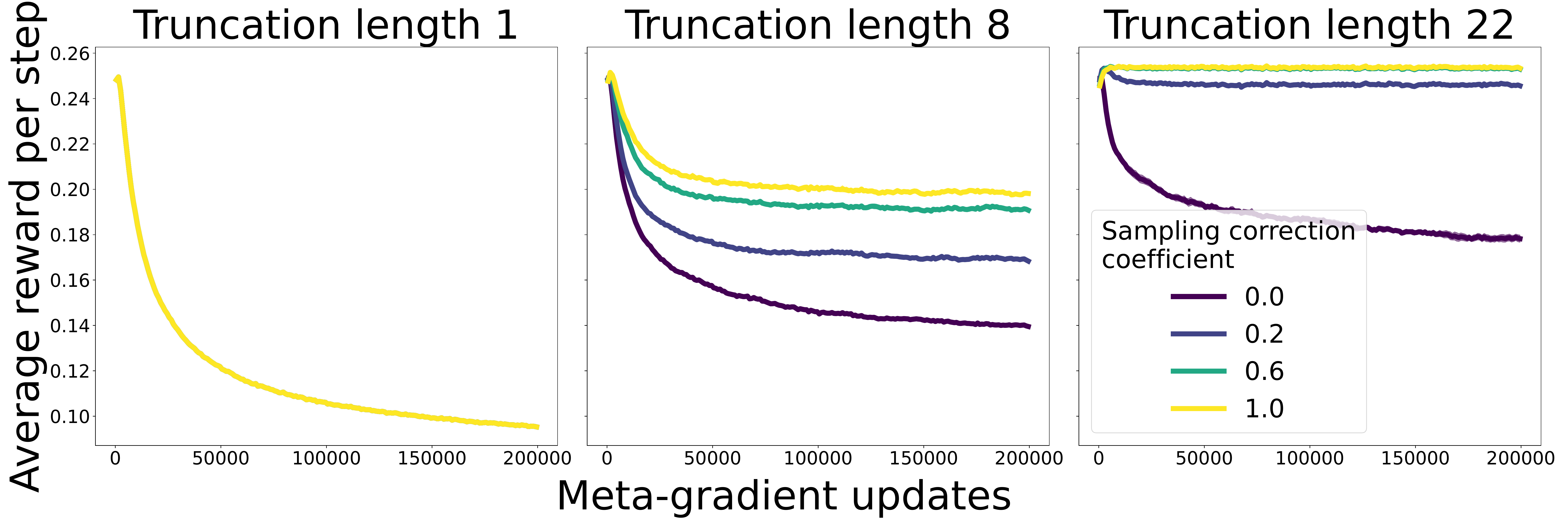}%
      \caption{Learning curves}
    \end{subfigure}
    \caption{\textbf{(a)} Comparing meta-gradient estimators in the bandit setting.
    In all panels, the x and y axes correspond to the two meta-parameters defining the learning rate schedule,
    the background shows the average return with the meta-parameter value sampled at the center of the cell,
    the crosses show the nine initial meta-parameter values,
    and the lines terminating in circles show the trajectories the parameters take during meta-training.
    \textbf{(b)} Meta-learning curves for the meta-gradient estimators in the bandit setting with truncated meta-optimization horizons.
    The shading of the curves shows the standard error across seeds.}
    \label{fig:bandit_curves}
  \end{figure}

We first investigate how the different meta-gradient estimators perform in an untruncated setting.
In \figref{fig:bandit_curves}, we compare four different meta-gradient estimators in the bandit setting by training the learning rate parameters starting from nine initializations.
The finite differences gradient estimator, which is unbiased, converges to the local optimum.
The naive estimator does not use the sampling correction, which leads to bias, causing the meta-parameters to move away from the local optimum.
The sampling corrected gradient estimator converges to the same local optimum as the finite differences one, confirming that it is unbiased.
We use DiCE \citep{foerster2018dice} to represent the category of meta-gradient estimators that use an estimate of the expected Hessian in the meta-gradient.
Since the bandit problem does not have a time dimension, some of the methods developed for reducing variance of the expected Hessian estimator, including those proposed by \citet{rothfuss2018promp} and \citet{farquhar2019loaded}, simply reduce to DiCE.
Using DiCE does not stop the meta-optimization from escaping the local optimum, demonstrating that the DiCE meta-gradient is biased.

\subsection{Bias-variance tradeoff of meta-gradient estimators}
\label{subsec:bias_variance_tradeoff}
\begin{wrapfigure}{r}{0.45\textwidth}
    \vspace*{-1em}
    \begin{center}
    \includegraphics[width=0.45\textwidth]{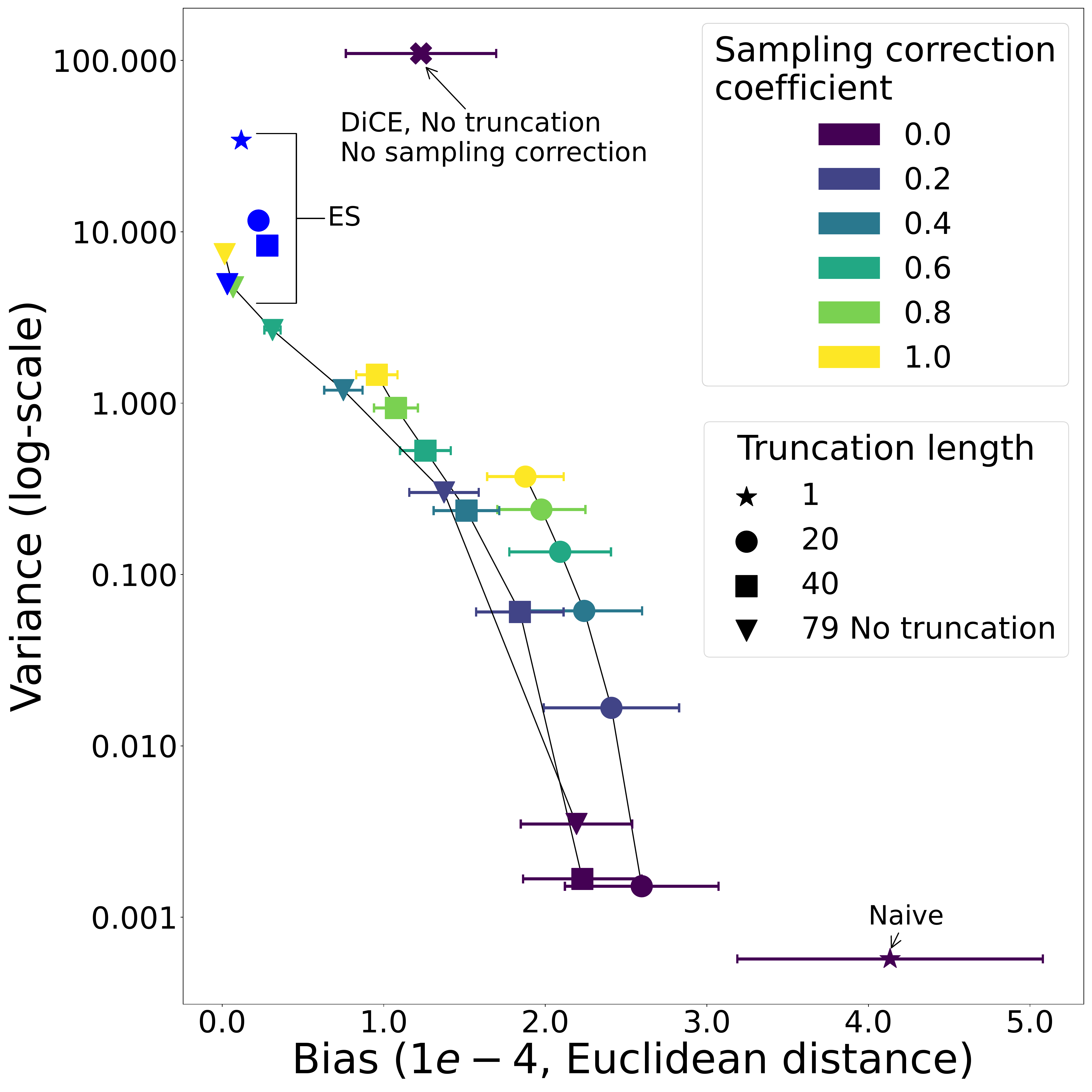}
    \end{center}
    \caption{
        The bias and variance estimates for different meta-gradient estimators in the bandit setting.
        The error bars show the standard deviation of the bias when computed across bootstrap samples of the initializations.
        The different markers correspond to different truncation lengths as given in the legend.
        The different colors correspond to different sampling correction coefficients.
    }
    \label{fig:bias_variance_frontier}
    \vspace*{-1em}
\end{wrapfigure}
Next, we investigate the bias-variance tradeoffs of the meta-gradient estimators.
We consider the tradeoffs with respect to the truncation length and the sampling correction coefficient $\lambda$.
We also show how estimators using DiCE and ES compare.
The truncated meta-gradient estimators are computed by randomly sampling a fixed-length window of inner loop updates during the lifetime and restricting backpropagation to that window.
For each point in the comparison, we sample a large number of such windows.
The objective of the agent is the same as before, i.e., the average lifetime return, which is approximated with the average return within the truncation window.

The results of the comparison are presented in figure~\ref{fig:bias_variance_frontier}.
The bias is estimated by computing the average Euclidean distance to the true meta-gradient (computed to high precision with finite differences) at 25 points around the optimum of a bandit problem similar to the one shown in figure \ref{fig:bandit_curves}.
The error bars show the standard deviation of the bias estimated with 10k bootstrap samples from the 25 points, to reflect how much the bias varies across the meta-optimisation landscape.
Because the sampling correction coefficient and truncation window length change the meta-gradient magnitude, while its direction may be more important in practice, we also present the same data in figure \ref{fig:bias_variance_cosine_similarity}
using cosine similarity as the bias measurement; the results are qualitatively similar.

\textbf{Effect of the truncated backpropagation.}
Figure \ref{fig:bias_variance_frontier}
shows that the general trend is for the bias to decrease with the increased truncation length.
However, for multiple truncation lengths, the standard deviation of the bias of the uncorrected estimator overlaps with the mean of the other truncation lengths.
This suggests that the bias of the uncorrected estimator does not decrease monotonically with the increased truncation length but may actually grow when a longer window is considered.


\textbf{Effect of the sampling correction coefficient.}
Figure~\ref{fig:bias_variance_frontier} confirms that the untruncated and sampling corrected meta-gradient estimator has no bias.
It also shows that increasing $\lambda$ tends to decrease bias.
However, we observe that for the truncated estimators, for some of the initializations the bias does not decrease with $\lambda$.
This is because the sampling correction terms themselves are biased in the truncated case and increasing their weight may increase the total bias at some parts of the meta-parameter space.
We demonstrate this effect in figure~\ref{fig:growing_bias_with_sampling_correction} in the appendix, which shows a case where the bias increases with $\lambda$ for one initialization and decreases for another.

\textbf{Variance of the estimators.}
The variance of the meta-gradient estimators grow rapidly with $\lambda$ and the truncation length.
The variance of the unbiased estimator ($\lambda = 1$, truncation length = 79) is multiple orders of magnitude higher than that of the naive estimator ($\lambda = 0$, truncation length = 1).
This is partly explained by the naive estimator having an order of magnitude smaller norm than the true gradient.
In settings with long truncation lengths and large inner-loop batch sizes, the high variance from the sampling correction may prevent its use without more extreme variance reduction methods.

\textbf{Evolution strategies.}
While ES has high variance, figure~\ref{fig:bias_variance_frontier} shows a case where its variance is smaller than that of the unbiased backpropagation-based estimator.
This is possible first because the black box design does not require an analogue to the sampling correction to account for the effect of the meta-parameters on the data distribution, and second because the variance of ES does not grow with the truncation length as rapidly as that of backpropagation based meta-gradients.
In this case, the truncated ES estimators have larger variance than the untruncated one, because the truncated ones are averaged over random windows along the lifetime.

\textbf{Pareto frontier.}
The bias-variance tradeoff presented in figure~\ref{fig:bias_variance_frontier} gives rise to a Pareto frontier.
Only two points are guaranteed to be on the frontier.
The ``naive'' estimator with truncation length 1 is on the frontier because it has the lowest variance out of all of the estimators considered.
The other point is the untruncated and sampling-corrected backpropagation-based estimator, which is the only point that is guaranteed to be unbiased.
However, in practice, the bias from the smoothing of the objective in ES can be small and ES can therefore be on the frontier too depending on the specifics of the problem, especially the length of the lifetime.
In general, ES has high variance, but its variance has a more favorable dependence on the length of the lifetime than the backpropagation-based estimators, so for very long lifetimes it will have lower variance than the backpropagation-based counterpart.
While the other estimators are not guaranteed to be on the frontier, in the problems we consider we see that, increasing the truncation length has a favorable bias-variance tradeoff to increasing sampling correction coefficient.

\subsection{Using sampling-corrected truncated estimators}
\label{sec:bandit_meta_learning}
In the previous experiment, we found that the bias of the truncated meta-gradient estimator may increase with the sampling correction coefficient under some conditions.
To investigate how using the sampling correction with the truncated estimator works in practice, we train the meta-parameters of the bandits with the different estimators.
We train the meta-parameters for 200k steps using Adam \citep{kingma2014adam}.
Figure \ref{fig:bandit_curves} suggests that higher sampling correction values asymptote to better local optima.
Thus, even though the sampling correction can increase the bias of truncated estimators, it can still lead to better meta-learning performance.


\subsection{Sampling corrections for the full RL problem}
\label{subsec:gridworld}
\begin{figure}
    \begin{center}
    \includegraphics[width=0.8\textwidth]{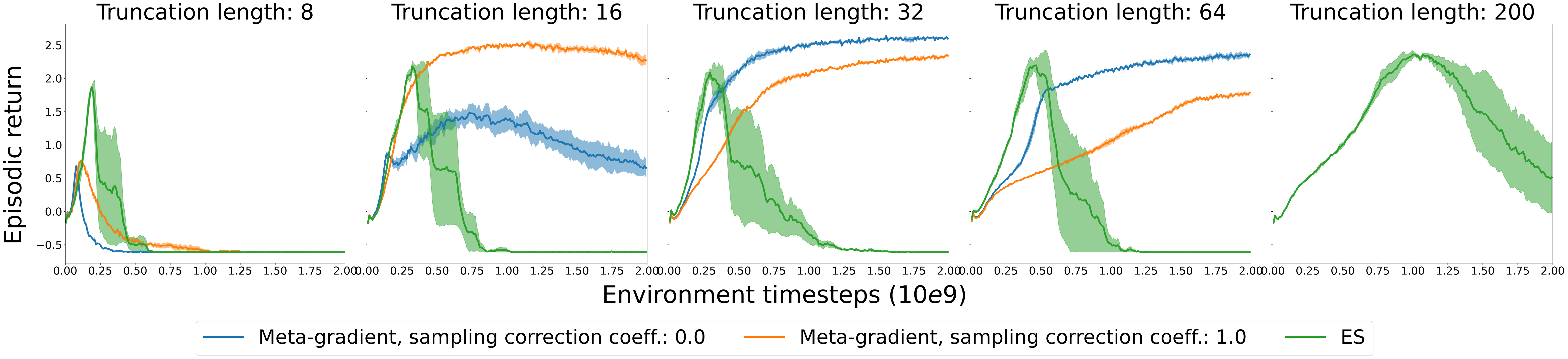}
    \end{center}
    \caption{
        Meta-learning curves for the meta-gradient estimators in the MDP setting with truncated meta-optimization horizons.
        The curves show a smoothed version of the episodic returns throughout the meta-learning.
        The shading of the curves shows the standard error across seeds.
    }
    \label{fig:mdp_learning_curves}
\end{figure}
We now compare some of the meta-gradient estimators on or close to the Pareto frontier in the full RL problem.
We consider a nonstationary gridworld environment adapted from \citet{flennerhag2021bootstrapped} where the agent searches for one of two rewarding objects.
The nonstationarity arises because the rewards of the objects are swapped every $6400$ steps.
Due to the nonstationarity, no unbiased gradient estimator exists.
Nevertheless, we expect the truncation horizon and the sampling correction to matter in this setting as well.
The agent is implemented as a neural network without memory.
The meta-learning problem is to learn a small neural network that outputs the entropy coefficient used for training the agent using a history of recent rewards as its input.
The inner-loop trains the policy with a standard actor-critic algorithm and updates the parameters using SGD.
The outer-loop updates the meta-parameters by gradient descent on the gradient given by equation~\ref{eq:sampled_meta_gradient}, or ES gradient estimate on~\ref{eq:meta_objective} and uses Adam for optimization.
For full details on the setting and hyperparameters see appendix \ref{appendix:mdp_details}.

The learning curves for training the entropy schedule with backpropagation-based and ES estimators are shown in figure~\ref{fig:mdp_learning_curves}.
As with the bandits, the truncation length has a significant impact on the meta-optimization, largely determining what kind of local optimum it converges to.
The sampling correction seems to lead to improved performance for the shorter truncation lengths, but results in very slow learning for the longer ones.
ES learns quickly but diverges from the local optimum it reaches, making its overall performance poor.
The meta-parametrization it learns is qualitatively different from those learned by the other estimators as can be seen from appendix~\ref{appendix:mdp_details}.
These results suggest that at least in this setting, the truncation length is the main driver of the meta-learning performance, sampling correction can be important in some cases, but ultimately using it makes the meta-learning very slow, and that achieving stable meta-learning with ES can be difficult.


\section{Related work}
The estimation of meta-gradients has been studied in the context of short-horizon multi-task meta-RL algorithms.
In MAML \citep{finn2017model}, the initialization of an agent is learned such that on a new task, its performance after a few policy gradient updates is maximized.
\citet{al2017continuous}
derive an unbiased meta-gradient estimator for learning the initialization and extend it to a continuous adaptation setting.
\citet{stadie2018some} provide a similar derivation and point to the improved exploratory behavior of the initial policy in MAML as an additional motivation for using the unbiased estimator.
\citet{fallah2020provably} study the convergence properties of the unbiased meta-gradient estimator.
In concurrent works, \citet{tang2021biased} and \citet{liu2021settling} also show that the DiCE-style Hessian estimator is a biased estimator of the true meta-gradient.
In addition to discussing the bias from the DiCE-style meta-gradient estimation,
we investigate the bias-variance tradeoff due to the sampling correction and truncation.

As confirmed by our experiments, long truncation lengths lead to high variance whereas short lengths increase bias.
\citet{bonnet2021one} propose to make the bias-variance tradeoff in the truncation length by mixing multiple truncation lenghts in a single update.
\citet{flennerhag2021bootstrapped} take another approach to the tradeoff by proposing a new bootstrapping meta-objective.
These works consider the bias-variance tradeoff due to truncated backpropagation, but they ignore the bias from missing the sampling correction terms, which we investigate together with the bias from truncation.

Besides RL, meta-gradients have been actively studied in supervised learning.
\citet{wu2018understanding} show that for some problems the bias from truncating the optimization horizon yields suboptimal solutions.
\citet{metz2019understanding} demonstrate how backpropagation through many update steps yields highly non-smooth objective surfaces making meta-learning challenging.
\citet{franceschi2017forward} show that for a small number of meta-parameters, forward-mode gradient computation can be used to overcome the memory constraint due to the long optimization horizon.
These works demonstrate how the optimization horizon can be important for meta-learning and propose methods to alleviate the problems preventing meta-optimization over long truncation horizons.
Despite their applications in supervised learning, these findings consider backpropagation through gradient-based updates and as such apply to meta-gradient estimation for RL as well.


\section{Conclusion and limitations}
In this paper, we investigated the estimation of meta-gradients in meta-RL.
We showed that, contrary to claims in prior work, using an estimator of the expected Hessian in the meta-gradient estimator always adds bias, and is likely to also increase variance.
Furthermore, we discussed the sampling correction needed for unbiased meta-gradient estimates, and described how it may be employed in the truncated optimization setting.
In doing so, we explored the space of bias-variance tradeoffs that can be made in meta-gradient estimation.
We also considered using ES instead of backpropagation for estimating meta-gradients.
While ES can have lower variance than backpropagation for long horizons, we were unable to achieve stable meta-learning using it.
We found all of the estimators considered to have severe drawbacks: ES can be unstable, backpropagation scales poorly with the truncation length and is biased without sampling correction, which in turn results in even higher variance.
Therefore, much work remains in developing better meta-gradient estimators that are capable of training complex meta-parametrizations on complicated problems.

The empirical analysis presented in this paper is limited to only a small number of relatively simple problems, and is produced with finite resources for hyperparameter search.
While we kept things simple in order to avoid introducing confounders, it is possible that the empirical results do not generalize to all other domains.
As for negative societal impacts, our research is unlikely to have any directly but any progress in meta-RL can lead to negative externalities through applications.





\begin{ack}
Risto Vuorio is supported by EPSRC Doctoral Training Partnership Scholarship and Department of Computer Science Scholarship.
Jacob Beck is supported by the Oxford-Google DeepMind Doctoral Scholarship.
This work was supported by a generous equipment grant from NVIDIA.
\end{ack}

\bibliography{references}
\bibliographystyle{icml2022}

\appendix

\section{Derivation of the unbiased meta-gradient}
\label{appendix:meta_grad_derivation}
The unbiased meta-gradient estimator can be derived as follows
\begin{align}
    \hspace{-0.1em}\nabla_{\eta} J_K(\eta) &= \nabla_{\eta} \sum_{k=0}^{K} \E_{\substack{\{\D^i \sim p(\D^i ; \theta^{i})\}_{i=0}^{k-1}}} \bigg[ \E_{\tau \sim p(\tau ; \theta^k)} \left[R(\tau)\right]\bigg] \\
    &= \nabla_{\eta} \sum_{k=0}^{K} \int R(\tau) p(\tau; \theta^k) \prod_{i=0}^{k-1} p(\D^i ; \theta^i) d \D^i d\tau \\
    &= \sum_{k=0}^{K} \int R(\tau) \bigg(\nabla_{\eta} \theta^k \nabla_{\theta^{k}} \log p(\tau ; \theta^k)
     \nonumber \\
     & \hspace{5em} + \sum_{j=0}^{k-1} \nabla_{\eta} \theta^j \nabla_{\theta^{j}} \log p(\D^j; \theta^j) \bigg) p(\tau ; \theta^k) \prod_{i=0}^{k-1} p(\D^i ; \theta^i) d \D^j d\tau\\
    &= \sum_{k=0}^{K} \E_{\substack{\{\D^i\}_{i=0}^{k-1} \\ \tau \sim p(\tau ; \theta^k)}} \bigg[ \bigg( \underdescribe{ \sum_{j=0}^{k - 1} \nabla_{\eta} \theta^j \nabla_{\theta^{j}} \log p(\D^j; \theta^j)}{\texttt{sampling correction}} + \underdescribe{\vphantom{\sum_{j=0}^{k-1}}\nabla_{\eta} \theta^k \nabla_{\theta^{k}} \log p(\tau ; \theta^k)}{\texttt{direct meta-gradient}} \bigg) R(\tau)\bigg]. \label{eq:unbiased_meta_grad_full_derivation}
\end{align}

\section{Meta-gradient estimators for expected policy gradients}
\label{appendix:expected_meta_gradient}
An expression for the meta-gradient without the sampling correction, used by most meta-gradient algorithms, can be derived by substituting the stochastic policy gradient in the update function with the expected policy gradient given by \eqref{eq:expected_policy_gradient}.
In that case, there is no dependency between the policy $\theta^k$ and the data sampled with earlier policies because when the policy is updated using the expected policy gradient, $\theta^k$ becomes a deterministic function of $\eta$ and $\theta^0$.
Therefore, we no longer get the sampling correction terms that consider how $\eta$ affects the distribution of data.
The meta-gradient for the expected update case is given by
\begin{align}
    \nabla{\eta} J_{K}'(\eta) = \nabla_{\eta} \sum_{k=0}^{K} \E_{\tau \sim p(\tau ; \theta^k)} \bigg[R(\tau)\bigg]
    =  \sum_{k=0}^{K} \E_{\tau \sim p(\tau ; \theta^k)} \bigg[ \nabla_{\eta} \theta^k \nabla_{\theta^{k}} \log p(\tau; \theta^k) R(\tau)\bigg]. \label{eq:expected_inner_meta_grad}
\end{align}
Algorithms that estimate this by computing inner-loop updates and the meta-gradient from samples are biased because they ignore the sampling correction terms.

\section{Exponential discounting of sampling corrections for variance reduction}
\label{appendix:exp_discount_meta_grad}
We consider an alternative weighting using exponential discounting motivated by analogy to the usual discounting procedure in RL.
The exponentially discounted meta-gradient estimator is expressed as follows
\begin{align}
    \sum_{k=0}^{K}
     \frac{1}{|\D|} \bigg( \sum_{j=0}^{k - 1} \alpha^{k - j} \nabla_{\eta} \theta^j \nabla_{\theta^{j}} \log p(\D^j ; \theta^j)  \sum_{\tau \in \D^k} R(\tau)
      +  \sum_{\tau \in \D^k}  \nabla_{\eta} \theta^k \nabla_{\theta^{k}} \log p(\tau ; \theta^k) R(\tau) \bigg), \label{eq:exp_discount_meta_grad}
\end{align}
where $\alpha$ is a meta-discount factor.
Analogously to discounting in RL, the weights on the sampling correction terms become exponentially smaller the further back along the update trajectory they are from the return at $k$.
Unlike in standard RL, the discount on the sampling correction term uniformly weights all of the terms in a batch, because all the data in the batch arrives to the update $\Psi(\eta, \theta, \D)$ simultaneously.

We compare this variance reduction scheme with the uniform weighting experimentally.
In figure \ref{fig:meta_discount},
we show that the exponential discounting of the sampling correction terms results in a similar bias-variance tradeoff to the simpler sampling correction coefficient.
Therefore, at least in this bandit setting, exponential discounting does not seem to have an advantage over using the uniform weighting.

\section{Advantage estimation with sampling corrections}
\label{sec:advantage_experiment}
\begin{figure}
    \centering
    \includegraphics[width=0.5\textwidth]{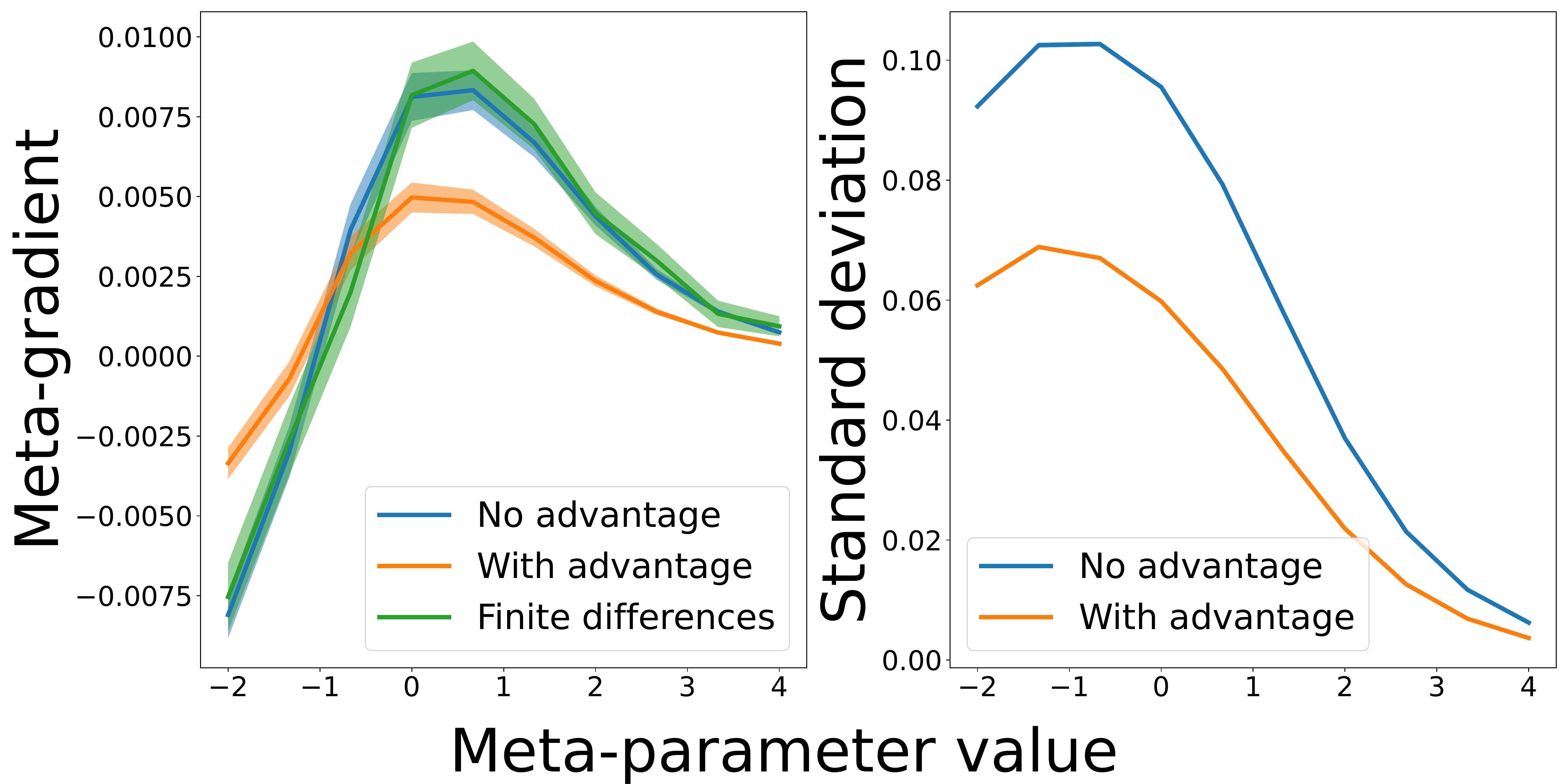}
    \caption{
    Comparing meta-gradient estimators with and without advantage estimation with a ground truth computed by finite differences.
    The x-axis is the value of the entropy coefficient being tuned with meta-gradients.
    In the left panel, the meta-gradient is computed at different meta-parameter values.
    The shading shows $95\%$ confidence interval of the mean.
    In the right panel, the standard deviation of the meta-gradient is computed at different meta-parameter values.
    The bootstrap estimate of the $95\%$ CI for the standard deviation estimate is sufficiently small that it cannot be rendered in this figure.
    } \label{fig:advantage_bias_variance}
\end{figure}
In the previous experiment, the variance of the meta-gradient was mitigated by using a large batch size to estimate the meta-gradient.
In more complex RL problems, variance reduction via algorithmic means is important due to limited computation budgets making it difficult to use large batch sizes.
In this experiment, we demonstrate that using a standard advantage estimator with the sampling correction results in bias and illustrate the resulting bias-variance tradeoff.
This experiment considers a similar problem setting as the previous one with the main differences that the nonstationarity is removed by only considering a single truncation window starting from the initial parameters and that a single scalar entropy coefficient is tuned with the meta-gradient.
Sigmoid activation is applied to the entropy coefficient.
Details and hyperparameters of the experiment are provided in appendix~\ref{appendix:mdp_details}.

Figure \ref{fig:advantage_bias_variance} shows the bias and variance of the untruncated, sampling corrected meta-gradient estimator with and without advantage estimation.
The bias is computed as the Euclidean distance to a finite differences estimate of the meta-gradient.
The comparison verifies the results in section \ref{sec:advantage}: using the advantage estimator with the sampling correction does reduce variance, but unlike with standard policy gradients, it introduces bias.
In practice, whether trading off variance for bias is practical depends on the problem setting.
In our experiments, we chose not to use the advantage estimators to focus on the bias from truncation and sampling correction, but using the advantage estimator may be worthwhile in practice.
In a practical meta-gradient algorithm using sampling corrections, variance reduction is crucial for performance, so research into unbiased advantage estimators in this context is a promising direction for future work.

\section{Bandit settings}
\label{appendix:bandit_details}
In this section we describe the details of the bandit experiments.
The experiments are conducted in a multi-armed bandit setting.
The bandits have 30 arms with the arm means sampled from $exp(Uniform(-100, 1))$.
The reward for each pull is computed by adding noise sampled from Gaussian with standard deviation 2 to the arm mean.
The inner learning problem is learning the policy for the bandit and the outer problem is to learn a learning rate schedule for the inner learner.
The learning rates are grouped into two buckets to produce a two-dimensional meta-parameter for the ease of visualization.
The first eight learning rates are grouped into one bucket and the remaining twenty one into another, choosing an uneven split because the first few updates have larger impact on the final performance than the later ones.

The policy is parametrized as a softmax over a randomly initialized vector of logits.
The inner loop uses a simple policy gradient algorithm \citep{williams1992simple}.
The inner loop updates the policy for 29 update steps sampling 30 batches of experience in total.
Each update in the inner loop is computed on ten samples from the bandit.
The outer loop updates the learning rate schedule with the meta-gradient computed by \eqref{eq:sampled_meta_gradient}.
When truncation is applied, the inner loop is run as before but the outer loop only considers returns within the truncation window and only backpropagates gradients within the window.
The meta-gradient is computed across multiple inner learning problems in parallel.
For computing one meta-gradient update with the untruncated inner loop, inner batch size $\times$ lifetime length $\times$ parallel runs samples from the bandit are used.
All the updates are computed with stochastic gradient descent.
The hyperparameters of the experiments in each figure are summarized in Table \ref{table:hyperparameters}.
\begin{table}
\centering
\begin{tabular}{ccc}
& Hyperparameter & Value \\ \hline
\multirow{2}{*}{Shared hyperparameters} & inner batch size & 10 \\
& optimizer  & SGD \\ \hline
\multirow{3}{*}{Figure \ref{fig:bandit_curves} left panel} & parallel runs & 1000 \\
& lifetime length  & 30 \\
& outer loop updates & 100000 \\ \hline
\multirow{5}{*}{Figure \ref{fig:bias_variance_frontier} right panel  and figure \ref{fig:bias_variance_cosine_similarity}} & parallel runs & 1000 \\
& outer loop updates & 100000 \\
& outer learning rate & 0.01 \\
& lifetime length  & 80 \\
& number of random seeds & 10 \\ \hline
\multirow{5}{*}{Figure \ref{fig:bandit_curves} right panel} & parallel runs & 1000 \\
& outer loop updates & 200000 \\
& outer learning rate & 0.001 \\
& number of random seeds & 5 \\
& lifetime length  & 30 \\
& optimizer & ADAM \\ \hline
\multirow{5}{*}{Figure \ref{fig:truncated_learning_curves}, truncation lengths 1 and 8} & parallel runs & 1000 \\
& outer loop updates & 500000 \\
& outer learning rate & 0.05 \\
& lifetime length  & 30 \\
& number of random seeds & 5 \\ \hline
\multirow{5}{*}{Figure \ref{fig:truncated_learning_curves}, truncation lengths 22 and 29} & parallel runs & 1000 \\
& outer loop updates & 500000 \\
& outer learning rate & 0.005 \\
& lifetime length  & 30 \\
& number of random seeds & 5 \\ \hline
\multirow{3}{*}{Figure \ref{fig:bias_variance_frontier} left panel} & parallel runs & 2000 \\ 
& lifetime length  & 30 \\
& outer loop samples & 1000 \\ \hline
\end{tabular}
\caption{Bandit hyperparameters for each experiment \label{table:hyperparameters}}
\end{table}

\section{MDP settings}
\label{appendix:mdp_details}
In this section we describe the details of the gridworld experiments.
This experiment is inspired by the experiment in \cite{flennerhag2021bootstrapped}, but differs in details which are explained below.
The experiments are conducted in gridworld environment, which consists of a $5 \times 5$ room.
The room has two objects.
The task of the agent is to learn to navigate to the rewarding object repeatedly.
The positions of the agent and the objects are randomized in the beginning of the episode.
Each time the agent reaches one of the objects, the locations of the objects are randomized.
The environment is episodic with episode length capped at $16$ to avoid having to use value function for bootstrapping.
The agent receives a reward of $1$ for moving to a cell with one object and $-1$ for the other object.
Every timestep when the agent does not hit a rewarding object, it gets a reward of $-0.04$.
The action space of the agent consists of the cardinal directions.
When the agent hits the edge of the room, it stays in the same grid cell.
The observation space is the coordinates of the agent and the two objects represented as one-hot vectors, and an additional one-hot vector representing the timestep in the episode to make the environment fully observable.
The environment is made non-stationary by flipping the rewards of the objects every $6400$ timesteps.

The agent is trained using a simple policy gradient algorithm with a baseline trained alongside the agent and an entropy regularization term, which has its coefficient predicted by the meta-learner.
The agent is parametrized as an MLP with two layers of $256$ units and ReLU activations.
The value function is approximated by another MLP with the same architecture.
The value function is only used in the inner-loop.
The inner-loop computes updates over five environments in parallel and samples a full episode of $16$ steps from each of the environments for each update.
The parameters of the policy and value function are updated with SGD.
The meta-learning problem is to learn a small MLP, which outputs the coefficient for the entropy regularization term used by the inner-loop.
The meta-learner is parametrized as an MLP with a single hidden layer of $32$ units with ReLU activation.
The output activation is sigmoid.
The input to the MLP is a vector of mean rewards in the previous $10$ batches used for the inner loop update.
The meta-learner is updated by the meta-gradient given by \eqref{eq:sampled_meta_gradient} computed as an average across $50$ independent inner-loop learners running in parallel.
ADAM optimizer is used for the meta-learning.
Hyperparameters of the inner and outer-loops are given in table \ref{table:mdp_table}.

In section \ref{sec:advantage_experiment} we compare the bias and variance of the sampling corrected, untruncated estimator with and without advantage estimation.
The experiment is run in a maze setting similar to the previous one.
The agent parameters $\theta$ are reset at every reward flip interval, removing the nonstationarity from the point of view of the inner-loop.
The hyperparameters for the bias variance experiment are given in table \ref{table:mdp_table}.

\begin{table}
    \centering
    \begin{tabular}{ccc}
    & Hyperparameter & Value \\ \hline
    \multirow{4}{*}{Shared hyperparameters} & inner learning rate & 1.0 \\
    & inner value loss coefficient & 0.1 \\
    & inner optimizer & SGD \\
    & discount rate $\gamma$ & 0.99 \\
    \hline
    \multirow{9}{*}{Hyperparameters for \figref{fig:mdp_learning_curves}} & inner batch size & 5 \\
    & outer batch size & 50 \\
    & outer optimizer & ADAM \\
    & sampling horizon & 16 \\
    & maze size & $5 \times 5$ \\
    & reward flip interval & 6400 \\
    & sampling correction coefficient & $\{0.0, 1.0\}$ \\
    & truncation length $K$ & $\{8, 16, 32, 64\}$ \\
    & outer learning rate & $\{1e-6, 5e-6\}$ \\
    & number of random seeds & 3 \\
    & ES optimizer & ADAM as above \\
    & ES other hyperparameters & default evosax OpenES \\
    \hline
    \multirow{9}{*}{Hyperparameters for \figref{fig:advantage_bias_variance}} & inner batch size & 10 \\
    & outer batch size & 25 \\
    & outer optimizer & N/A \\
    & sampling horizon & 8 \\
    & maze size & $3 \times 3$ \\
    & reward flip interval & 64 \\
    & sampling correction coefficient & $1.0$ \\
    & truncation length $K$ & $8$ \\
    & outer learning rate & N/A \\
    & number of meta-gradient samples & 62500 \\
    \end{tabular}
    \caption{MDP hyperparameters for each experiment \label{table:mdp_table}}
\end{table}

\section{Source code release}
The source code for reproducing the results in this paper is provided at \url{https://github.com/vuoristo/meta-gradients}.

\section{Additional experimental results}
\label{appendix:exp_results}
\begin{figure}[h]
    \begin{center}
    \includegraphics[width=0.9\textwidth]{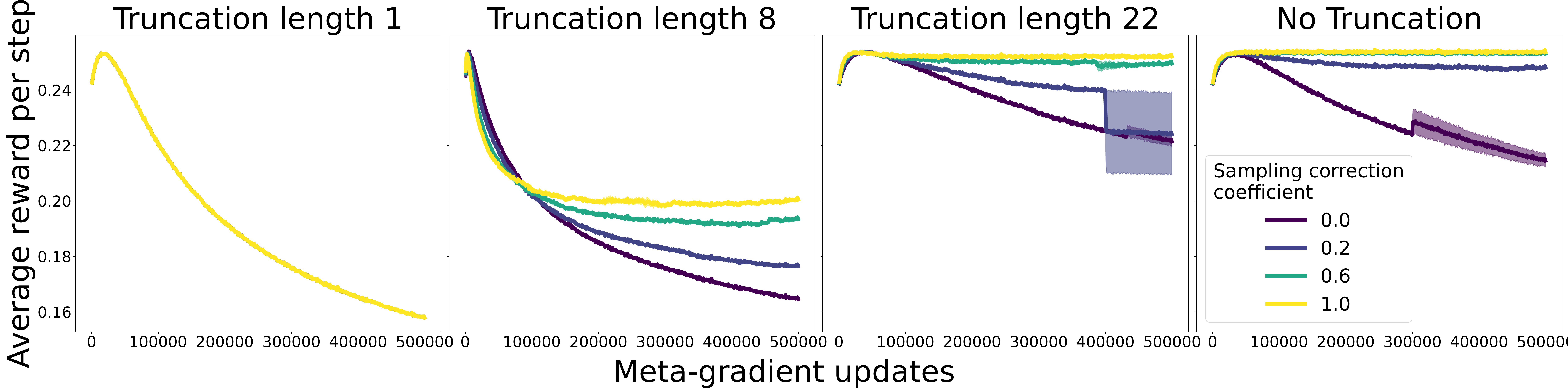}
    \end{center}
    \caption{
        Learning curves for the meta-gradient estimators in the bandit setting with truncated meta-optimization horizons using SGD as the optimizer.
        The shading of the curves shows the standard error across random seeds.
    }
    \label{fig:truncated_learning_curves}
\end{figure}
\begin{figure}[h]
    \begin{center}
    \includegraphics[width=0.9\textwidth]{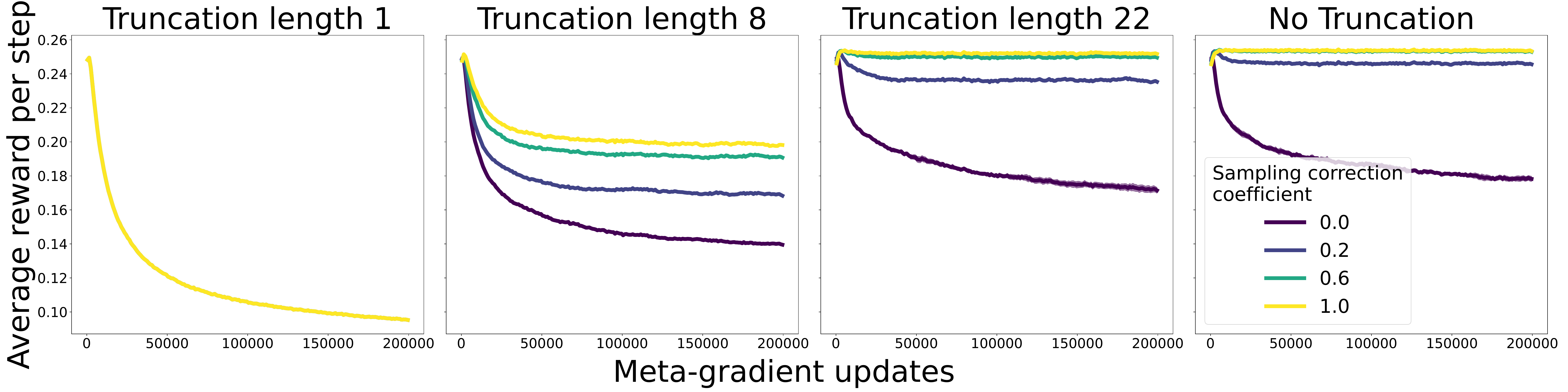}
    \end{center}
    \caption{
        Learning curves for the meta-gradient estimators in the bandit setting with truncated meta-optimization horizons using ADAM as the optimizer.
        The shading of the curves shows the standard error across random seeds.
    }
    \label{fig:truncated_learning_curves_adam_full}
\end{figure}
\begin{figure}[h]
    \begin{center}
    \includegraphics[width=0.45\textwidth]{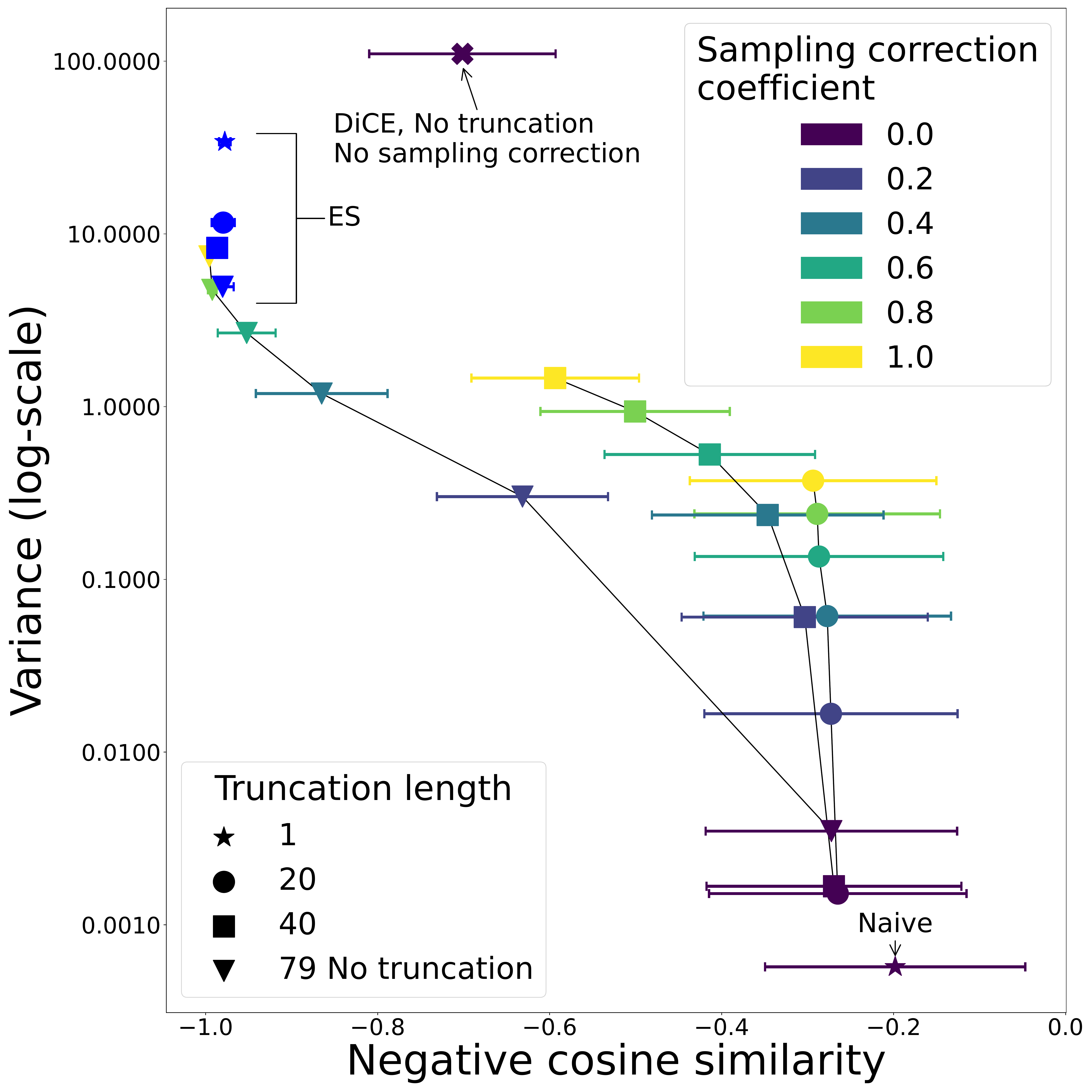}
    \end{center}
    \caption{
        The bias and variance of meta-gradient estimators in the bandit setting with negative cosine similarity estimate of the bias.
        Negative cosine similarity is shown instead of cosine similarity to keep the bias axis consistent with figure \ref{fig:bias_variance_frontier}, that is, bias grows toward the right edge of the figure.
    }
    \label{fig:bias_variance_cosine_similarity}
\end{figure}
\begin{figure}[h]
    \begin{center}
    \includegraphics[width=0.45\textwidth]{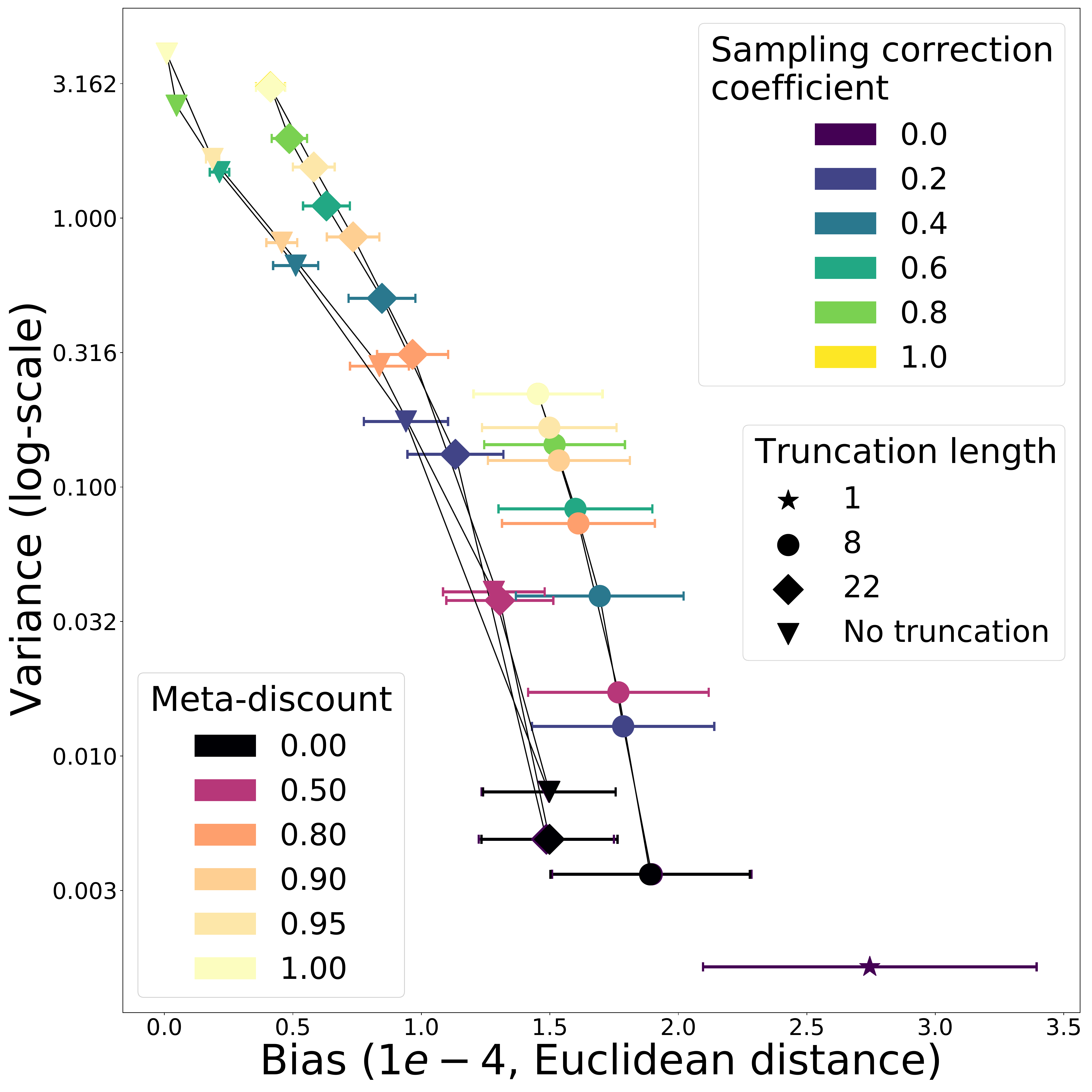}
    \end{center}
    \caption{
        The bias and variance of two different weighting schemes for the sampling correction terms are explored.
    }
    \label{fig:meta_discount}
\end{figure}
\begin{figure}
    \begin{center}
    \includegraphics[width=0.48\textwidth]{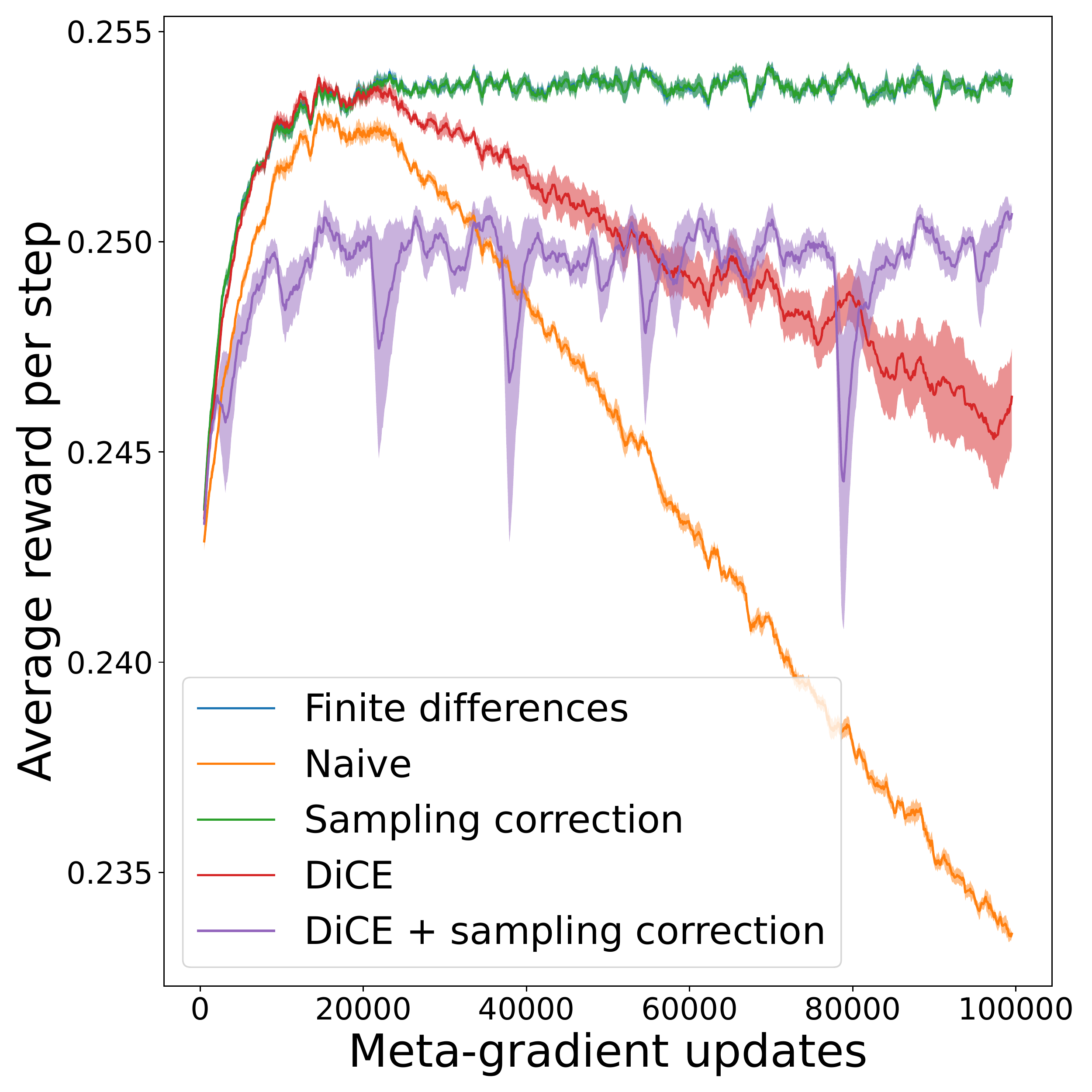}
    \end{center}
    \caption{
        Learning curves for the meta-gradient estimators in the bandit setting.
        The shading of the learning curves shows the standard error across random seeds.
        Note that the learning curves for the finite differences estimator and sampling corrected estimator are almost perfectly overlapped in this setting.
    }
    \label{fig:bandit_learning_curves}
\end{figure}
\begin{figure}
    \begin{center}
    \includegraphics[width=0.75\textwidth]{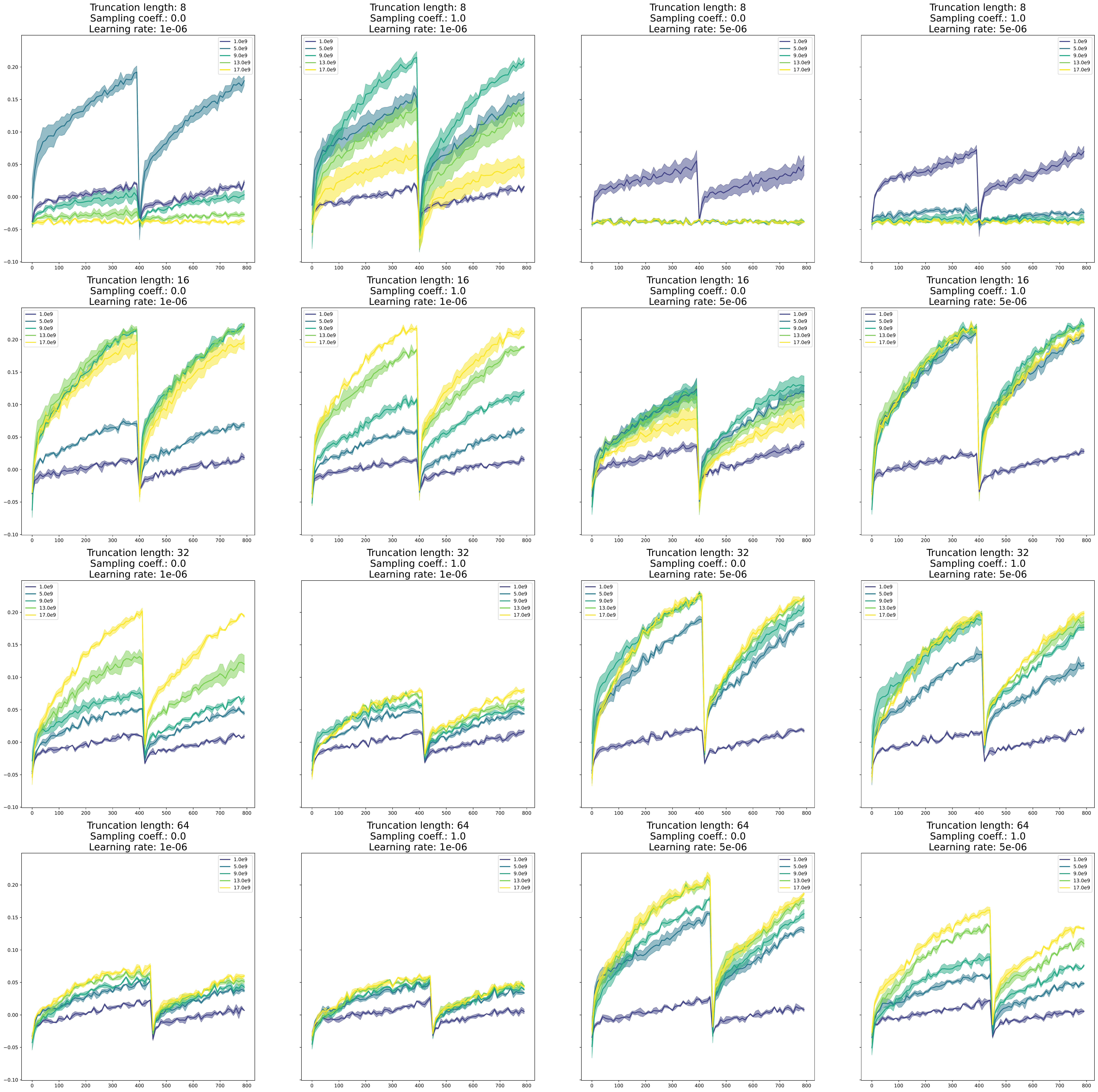}
    \end{center}
    \caption{
        Snapshots into the reward per timestep in the gridworld experiment.
        The figures are organized into columns by values of sampling correction coefficient and outer-loop learning rate and into rows by truncation lengths.
    }
    \label{fig:checkpoint_reward}
\end{figure}
\begin{figure}
    \begin{center}
    \includegraphics[width=0.75\textwidth]{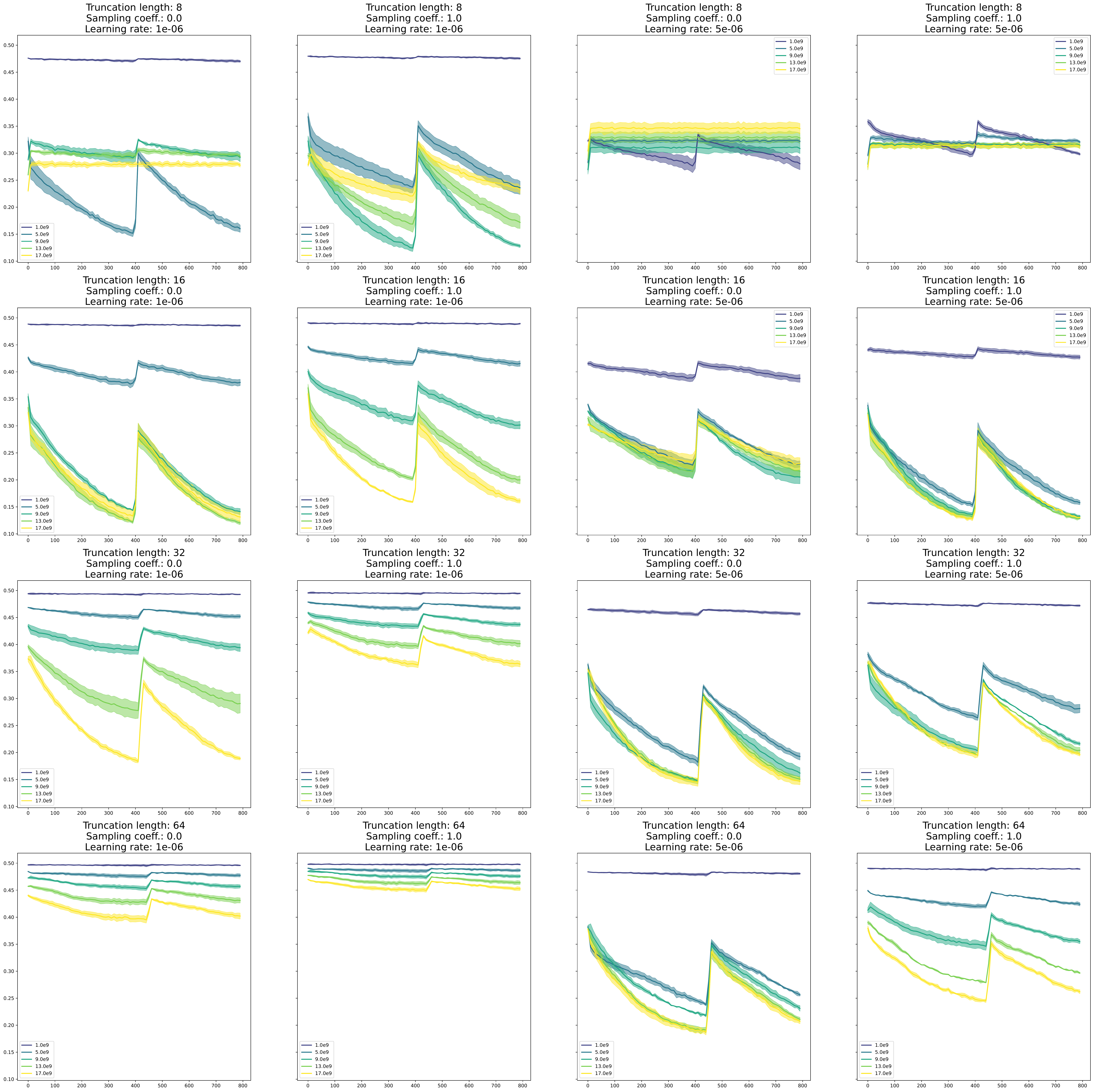}
    \end{center}
    \caption{
        Snapshots into the entropy regularizer coefficient per timestep in the gridworld experiment.
        The figures are organized into columns by values of sampling correction coefficient and outer-loop learning rate and into rows by truncation lengths.
    }
    \label{fig:checkpoint_ent_coef}
\end{figure}
\begin{figure}
    \begin{center}
    \includegraphics[width=0.75\textwidth]{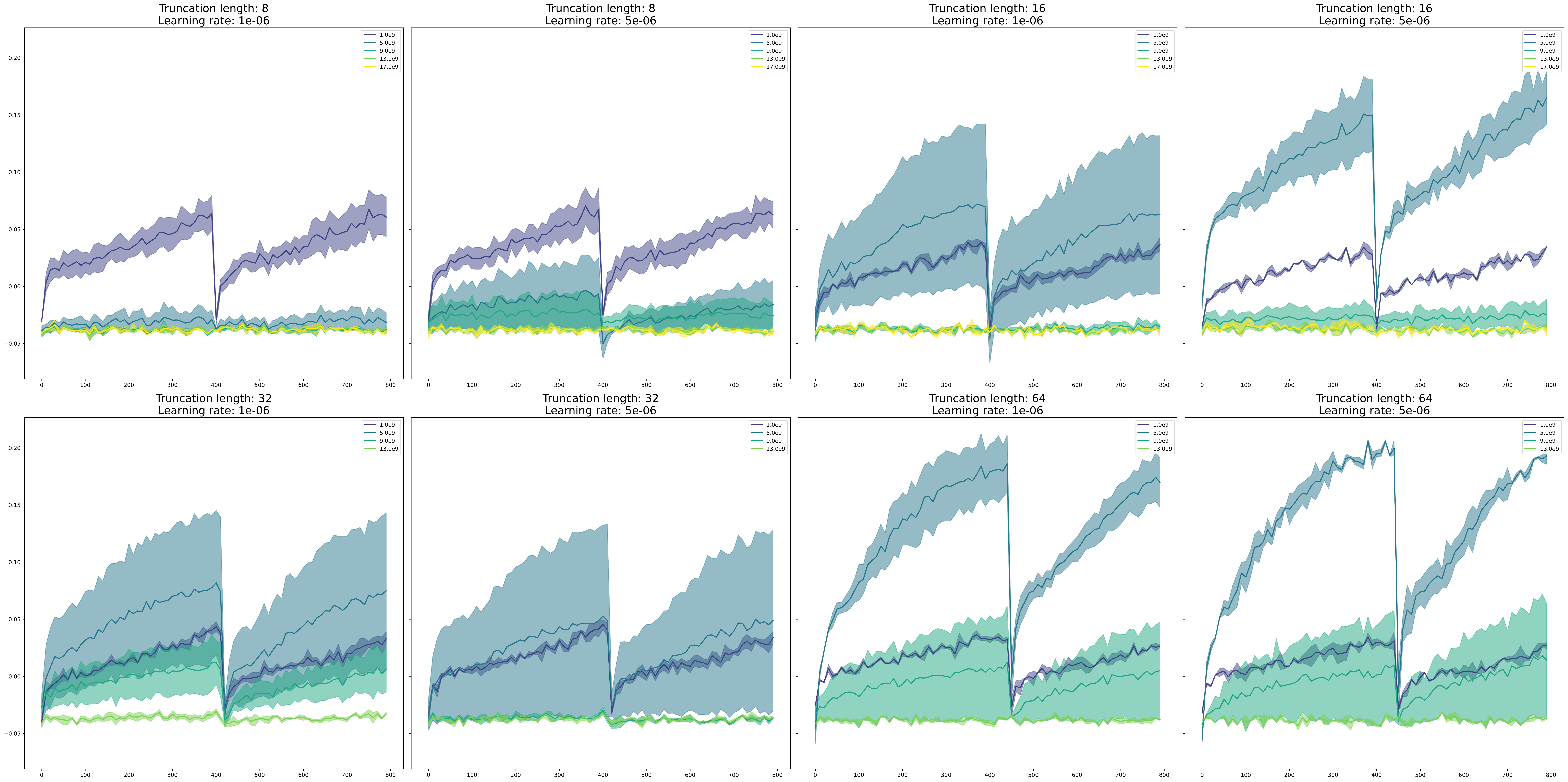}
    \end{center}
    \caption{
        Snapshots into the reward per timestep in the gridworld experiment with evolution strategies.
        The figures are organized into columns by values of sampling correction coefficient and outer-loop learning rate and into rows by truncation lengths.
    }
    \label{fig:checkpoint_reward_es}
\end{figure}
\begin{figure}
    \begin{center}
    \includegraphics[width=0.75\textwidth]{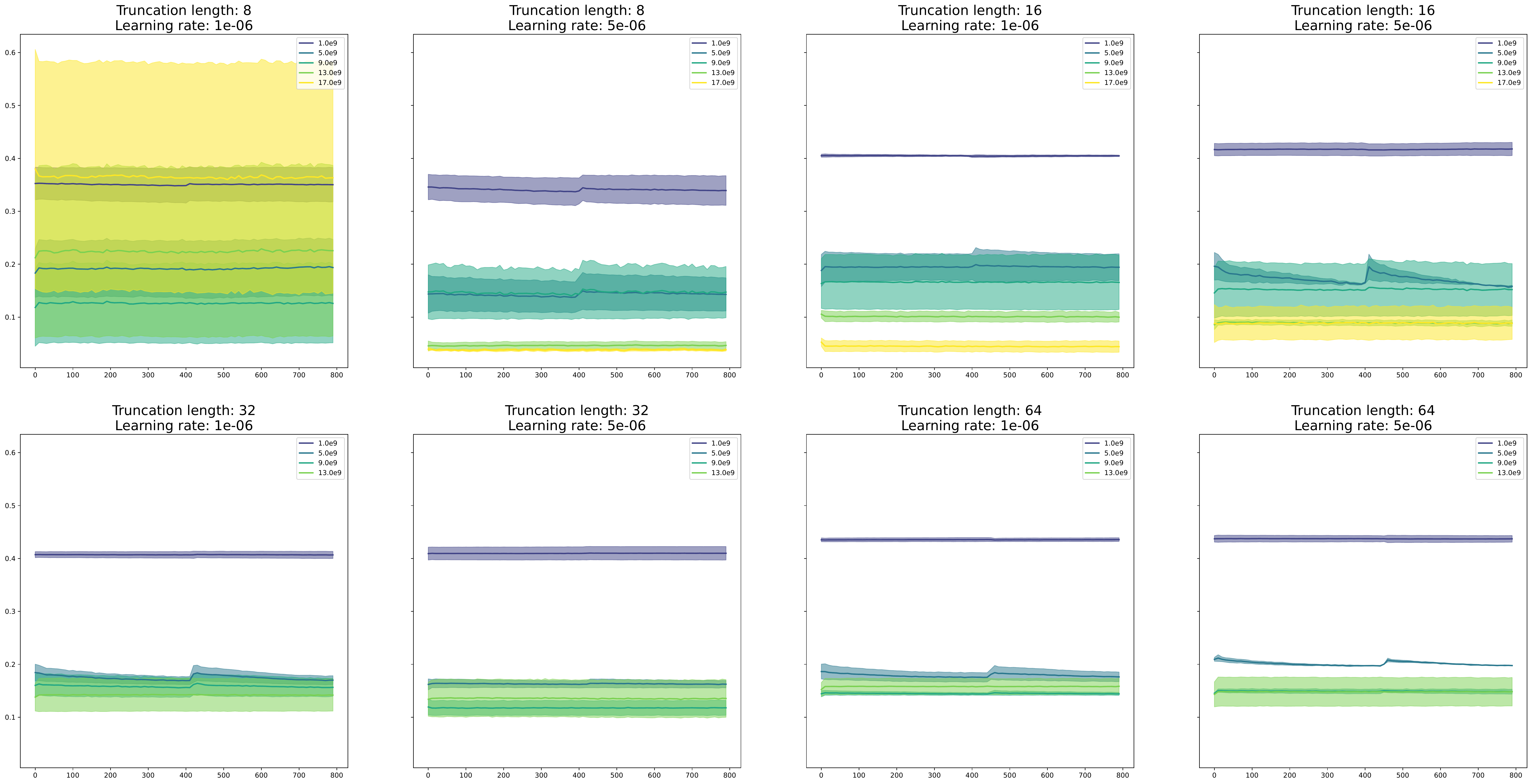}
    \end{center}
    \caption{
        Snapshots into the entropy regularizer coefficient per timestep in the gridworld experiment with evolution strategies.
        The figures are organized into columns by values of sampling correction coefficient and outer-loop learning rate and into rows by truncation lengths.
    }
    \label{fig:checkpoint_ent_coef_ess}
\end{figure}
\begin{figure}[h]
    \begin{center}
    \includegraphics[width=0.45\textwidth]{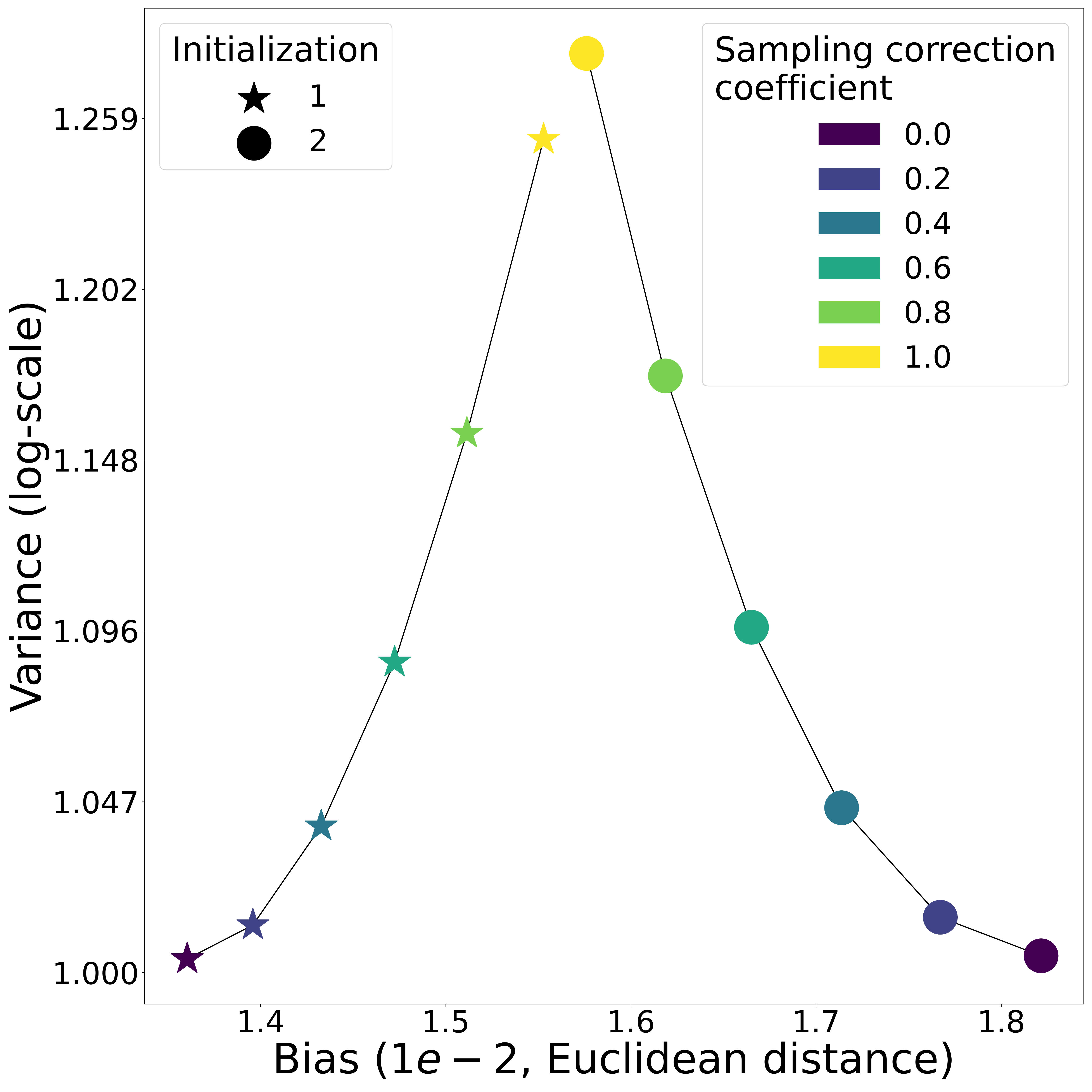}
    \end{center}
    \caption{
        Bias and variance for two initializations with a truncated estimator where the bias increases with the sampling correction coefficient for one initialization and decreases for the other
    }
    \label{fig:growing_bias_with_sampling_correction}
\end{figure}

The meta-gradient estimators are compared in terms of cosine similarity in the left panel of figure \ref{fig:bias_variance_cosine_similarity}.
The two weighting schemes for the sampling correction are compared in the right panel of figure \ref{fig:meta_discount}.
Learning curves for the truncated meta-gradient estimators using SGD as the optimizer are shown in figure \ref{fig:truncated_learning_curves}.
The truncated learning curves with ADAM as the optimizer are shown again in \figref{fig:truncated_learning_curves_adam_full}, with the added curves for truncation length 1.
The large error in the curves for the higher truncation lengths are due to some of the seeds becoming unstable during training and deviating far from the optimal region in a single gradient step.
This effect can be reduced in practice by using adaptive optimizers such as Adam and gradient clipping.
The learning curves with the different gradient estimators in the untruncated bandit setting are shown in \figref{fig:bandit_learning_curves}.
In the meta-learning curves, all algorithms start from the same initialization.
The gradients of the sampling corrected estimator are similar enough to the finite differences gradient in this setting that their respective learning curves are almost identical.
Figures \ref{fig:checkpoint_reward} and \ref{fig:checkpoint_ent_coef} show the reward per timestep and entropy coefficient respectively of the gridworld experiment in \ref{subsec:gridworld} evaluated at different snapshots during the meta-training.
The figures show two reward flip intervals.
Figure~\ref{fig:growing_bias_with_sampling_correction} shows bias and variance for two initializations with a truncated estimator where the bias increases with the sampling correction coefficient for one initialization and decreases for the other.

\end{document}